\documentclass[twocolumn]{style}

\usepackage[utf8]{inputenc} 
\usepackage{url}            
\usepackage{amsfonts}       
\usepackage{nicefrac}       

\usepackage{amsmath}
\usepackage{enumerate}
\usepackage{algorithm}
\usepackage{algorithmic}    
\usepackage{amsthm}
\usepackage{newtxtt}
\usepackage{diagbox}
\usepackage{colortbl}

\usepackage[framemethod=tikz]{mdframed}
\usepackage{amssymb}
\usepackage{bbm}            
\usepackage{xspace}
\usepackage{wrapfig}
\usepackage{adjustbox}
\usepackage{tabularx}
\usepackage{mathtools}

\usepackage{array}
\usepackage[normalem]{ulem} 
\usepackage{makecell}
\usepackage{enumitem}
\usepackage{float}
\usepackage{comment}

\usepackage{tikz}
\usetikzlibrary{arrows.meta,positioning,fit,calc,shapes.misc}
\usetikzlibrary{decorations.markings, backgrounds, shapes.geometric}

\definecolor{lumaBlue}{RGB}{92,36,131}     
\definecolor{lumaGreen}{RGB}{48,175,123}   
\definecolor{lumaRed}{RGB}{225,85,75}      
\definecolor{lumaGray}{RGB}{75,75,75}      
\definecolor{lightGray}{gray}{0.96}        
\definecolor{deepGray}{RGB}{50,50,50}      
\definecolor{lumaOrange}{RGB}{230,150,50}  

\definecolor{FrozenBlue}{RGB}{196,219,255}
\definecolor{TrainPeach}{RGB}{255,229,204}
\definecolor{EmbedGreen}{RGB}{197,224,180}
\definecolor{Ink}{RGB}{30,42,70}

\definecolor{OursGreen}{RGB}{232,245,233}
\definecolor{ColGray}{RGB}{245,245,245}
\definecolor{HeaderGray}{RGB}{250,250,250}

\definecolor{ablRow1}{RGB}{236,245,255}
\definecolor{ablRow2}{RGB}{221,238,255}
\definecolor{ablRow3}{RGB}{204,231,255}
\definecolor{ablRow4}{RGB}{187,224,255}
\definecolor{ablRow5}{RGB}{168,216,255}
\definecolor{ablRow8}{RGB}{130,200,255}

\tcbuselibrary{skins, breakable}
\tcbset{
  lumaBox/.style={
    enhanced,
    colback=lumaBlue!5!white,
    colframe=lumaBlue!60!black,
    fonttitle=\bfseries,
    coltitle=black,
    boxrule=0.5pt,
    arc=2mm,
    left=4pt,
    right=4pt,
    top=3pt,
    bottom=3pt,
    breakable,
  },
  lumaBoxGreen/.style={
    enhanced,
    colback=lumaGreen!5!white,
    colframe=lumaGreen!60!black,
    fonttitle=\bfseries,
    boxrule=0.5pt,
    arc=2mm,
    left=4pt,
    right=4pt,
    top=3pt,
    bottom=3pt,
    breakable,
  },
  lumaBoxRed/.style={
    enhanced,
    colback=lumaRed!5!white,
    colframe=lumaRed!60!black,
    fonttitle=\bfseries,
    boxrule=0.5pt,
    arc=2mm,
    left=4pt,
    right=4pt,
    top=3pt,
    bottom=3pt,
    breakable,
  },
}

\tikzset{
  >={Latex[length=2mm]},
  every node/.style={font=\sffamily\scriptsize,align=center},
  mod/.style={rounded corners=3pt, draw=Ink, very thick, inner sep=3pt},
  frozen/.style={mod, fill=FrozenBlue},
  train/.style={mod, fill=TrainPeach},
  embed/.style={mod, fill=EmbedGreen},
  dashedbox/.style={draw=Ink, thick, rounded corners=3pt, dashed, inner sep=5pt},
  circ/.style={circle, draw=Ink, fill=white, inner sep=0pt, minimum size=6pt, very thick},
  plus/.style={circle, draw=Ink, fill=white, inner sep=0pt, minimum size=7pt, very thick},
}

\theoremstyle{plain}
\newtheorem{theorem}{Theorem}[section]

\newtheorem{corollary}[theorem]{Corollary}
\theoremstyle{definition}

\theoremstyle{remark}
\newtheorem{remark}[theorem]{Remark}

\usepackage[textsize=tiny]{todonotes}

\DeclareMathOperator*{\argmin}{\mathop{\mathrm{argmin}}}

\newcommand{\camera}[1]{}                       

\title{Weight Space Representation Learning\\
via Neural Field Adaptation}

\author[1]{Zhuoqian Yang}
\author[1]{Mathieu Salzmann}
\author[1]{Sabine S\"usstrunk}

\affiliation[1]{EPFL}

\abstract{
We investigate the potential of weights to serve as effective representations, focusing on neural fields. Our key insight is that constraining the optimization space through a pre-trained base model and multiplicative low-rank adaptation (mLoRA) can induce structure in weight space. Across reconstruction, generation, and analysis tasks on 2D and 3D data, we find that mLoRA weights achieve high representation quality while exhibiting distinctiveness and semantic structure. When used with latent diffusion models, mLoRA weights enable higher-quality generation than existing weight-space methods.
}

\renewcommand{\takeawaysblock}{%
  \vskip 0.12cm
  {\small
    \setlength{\parskip}{0pt}%
    {\sffamily\bfseries Takeaways}\par
    \begin{itemize}[leftmargin=1.2em, itemsep=2pt, topsep=2pt, parsep=0pt]
        \item Properly constrained, symmetry-broken weights exhibit semantic structure and serve as effective data representations.
        \item \textbf{Structured weight space enables effective weight-space generation}.
    \end{itemize}
  }%
}

\date{\today}
\correspondence{\email{zhuoqian.yang@epfl.ch}}

\begin{document}


\newcommand*{\vertbar}{\rule[-0.25ex]{0.5pt}{1.5ex}}
\newcommand*{\horzbar}{\rule[.5ex]{2.5ex}{0.5pt}}
\newcommand{\dd}{\mathrm{d}}
\newcommand{\tkernel}{p}
\newcommand{\action}[2]{\left \langle #1, #2\right \rangle }
\newcommand{\bell}{\mathrm{b}}
\newcommand{\norm}[1]
{\left\Vert#1\right\Vert}
\newcommand{\Norm}[1]{\lvert \! \lvert \! \lvert #1 \rvert \! \rvert \! \rvert}
\newcommand{\abs}[1]{\left\vert#1\right\vert}
\newcommand{\babs}[1]{\Big \vert#1 \Big \vert}
\newcommand{\set}[1]{\left\{#1\right\}}
\newcommand{\parr}[1]{\left (#1\right )}
\newcommand{\brac}[1]{\left [#1\right ]}
\newcommand{\ip}[1]{\left \langle #1 \right \rangle }
\newcommand{\Real}{\mathbb R}
\newcommand{\Nat}{\mathbb N}
\newcommand{\Complex}{\mathbb C}
\newcommand{\eps}{\varepsilon}
\newcommand{\too}{\rightarrow}
\newcommand{\bbar}[1]{\overline{#1}}
\newcommand{\wt}[1]{\widetilde{#1}} 
\newcommand{\wh}[1]{\widehat{#1}} 
\newcommand{\diag}{\textrm{diag}} 
\newcommand{\dist}{d} 
\newcommand{\divv}{\mathrm{div}} 
\newcommand{\vol}{\mathrm{vol}} 
\newcommand{\snr}{\mathrm{snr}}
\newcommand{\logsnr}{\rho}
\newcommand{\trace}{\textrm{tr}} 
\def \bfi{\textbf{\footnotesize{i}}} 
\newcommand{\one}{\mathbf{1}}
\newcommand{\zero}{\mathbf{0}}
\newcommand{\vcc}[1]{\mathrm{vec}(#1)}
\newcommand{\mat}[1]{\bm{[} #1 \bm{]}}
\newcommand{\defe}{\coloneqq}

\def \etal{{et al}.}
\newcommand*{\eg}{{\it e.g.}\@\xspace}
\newcommand*{\ie}{{\it i.e.}\@\xspace}

\maketitle


\section{Introduction}
\label{sec:introduction}

Neural network weights have traditionally been viewed as opaque byproducts of optimization, high-dimensional vectors that encode learned functions but resist interpretation or manipulation. This perspective has begun to shift with recent advances in weight space learning, where researchers have demonstrated that network parameters can be merged \cite{yang2024modelmerging, singhhyperalign, lim2024empirical}, generated \cite{erkocc2023hyperdiffusion, peebles2023learning}, or used as inputs to other networks \cite{zhou2023nft, hospedales2020meta, lim2024learning}. A fundamental question nonetheless remains largely unexplored: Can neural network weights themselves serve as meaningful representations for data?

We investigate this question in the context of implicit neural representations (INRs), where neural networks are trained to overfit individual samples by mapping coordinates to values. INRs have proven to be versatile, capable of encoding diverse data modalities within a unified architecture \cite{xie2022neural}. Since INRs inherently encode signals as network parameters, using these weights as representations is a natural next step. However, neural network weights are known to be ambiguous by nature,
for example because neuron permutations and scaling can leave the network function unchanged \cite{zhao2025symmetry}; different random initializations 
may yield vastly different parameter configurations yet functionally identical models. In other words, functionally identical networks can be arbitrarily far in weight space \cite{zhao2025symmetry}, making the distribution multi-modal and difficult to learn, challenging the use of network weights as data representations.

\begin{figure}
    \centering
    \includegraphics[width=\columnwidth]{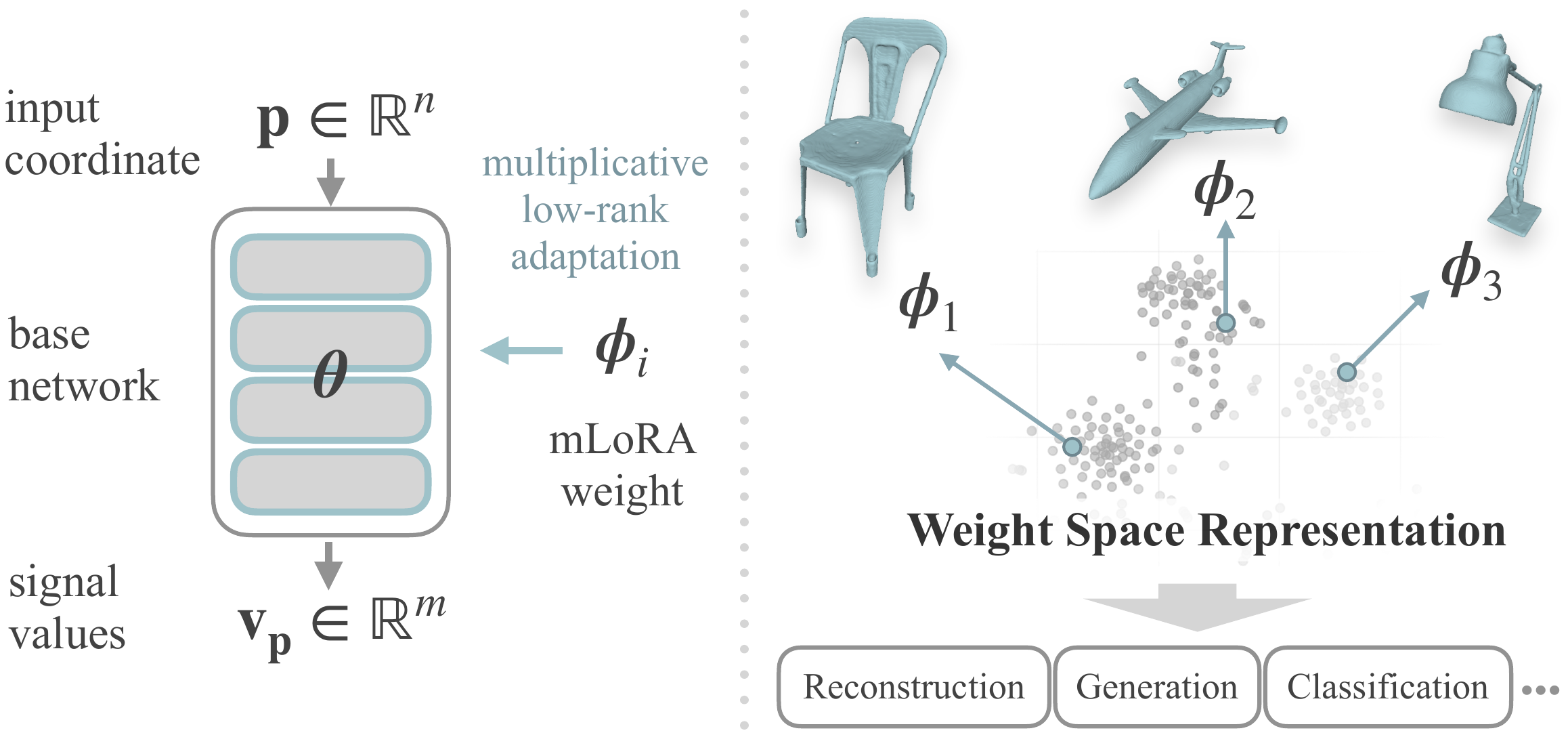}
    \caption{LoRA based weight space representation with neural fields. Given an input coordinate $\mathbf{p} \in \mathbb{R}^n$, a base neural field is adapted via mLoRA weights $\boldsymbol{\phi}_i$ to produce signal values $\mathbf{v}_\mathbf{p} \in \mathbb{R}^m$, each weight representing an instance. The mLoRA weights themselves form structured representations in weight space, enabling diverse applications.}
    \label{fig:paradigm}
\end{figure}

Our key insight is that constraining the network weights of different samples by introducing appropriate inductive biases allows us to 
transform these chaotic parameters into organized, semantic representations. To this end, as illustrated in Figure~\ref{fig:paradigm}, we propose to use Low-Rank Adaptation (LoRA) \cite{hu2022lora} within a pre-trained base neural field to create structured weight space representations. This design is motivated by two key properties of LoRA: First, adaptation through LoRA constrains the weight updates to lie within a low-dimensional subspace defined by the base model. Hu et al. \cite{hu2022lora} demonstrate this through subspace similarity analysis, showing that adaptations at different ranks share common singular vector directions, suggesting the existence of a meaningful low-rank adaptation subspace. Second, the low-rank constraint inherently reduces the dimensionality of the weight space representation, mitigating the curse of dimensionality that would otherwise hinder learning in high-dimensional parameter spaces.

We find that the standard additive LoRA formulation is insufficient for weight space learning in the context of neural fields. Instead, we introduce multiplicative LoRA (mLoRA), where weight updates are applied through element-wise multiplication rather than addition. This multiplicative formulation naturally aligns with modulation mechanisms in generative neural fields \cite{anokhin2021cips, karras2021alias, chan2020pigan}, where features are composed through multiplicative interactions, enabling effective weight space learning.

We validate our approach across multiple data modalities and tasks. First, we establish that LoRA-based weight representations achieve lower reconstruction error, greater consistency under different initializations, and better linear mode connectivity than standalone MLP weights. Second, we demonstrate that these structured weight spaces support generative modeling, with diffusion models trained on multiplicative LoRA weights outperforming prior attempts in weight space neural field generation. Finally, through evaluation on discriminative tasks (\textit{i.e.} classification and clustering), we confirm that the weight space structure correlates with semantic properties of the encoded data.

Our contributions can be summarized as follows:

\noindent 1. We demonstrate that independently optimized neural network weights, when properly constrained, 
can serve as effective data representations that capture semantic structure.

\noindent 2. We introduce multiplicative LoRA for neural fields, which we show provides superior representation quality compared to standard additive LoRA and standalone weight parameterizations.

\noindent 3. We validate weight space representations across diverse tasks: Reconstruction, generation, and classification, establishing their viability as a representation paradigm.

\section{Related Work}
\label{sec:related_works}

\subsection{Weight Space Learning}

The treatment of neural network weights as learnable representations has emerged as a distinct research direction, progressing from early hypernetwork approaches to sophisticated weight-space generative models. Early work demonstrated that network parameters could be generated by auxiliary networks~\cite{ha2016hypernetworks}, though these methods suffered from prohibitive memory overhead when scaling to modern architectures~\cite{wang2025scaling}. The fundamental challenge of weight space manipulation stems from permutation symmetry: Functionally identical networks can have vastly different parameter configurations due to neuron reordering~\cite{gao2024revisiting,navon2023equivariant}.

Recent advances have addressed these challenges through multiple strategies. Model merging techniques leverage optimal transport and activation matching to align neurons before parameter fusion~\cite{singh2020model,ormaniec2025fusion}, while equivariant architectures explicitly respect weight-space symmetries when processing network parameters~\cite{navon2023equivariant,kofinas2024graph}. A parallel research direction focuses on building neural networks that process weights as inputs. Neural Functional Transformers~\cite{zhou2023nft} and permutation-equivariant neural functionals~\cite{zhou2023permutation} construct architectures that can extract information from network parameters while respecting their symmetries. Methods operating on Low-Rank Adaptation (LoRA) weights~\cite{lim2024learning} develop GL-equivariant networks to process low-rank weight spaces of fine-tuned models. 
\camera{rephrased to highlight difference between ours and previous weight space learning}
Our work takes a different approach: rather than building model-agnostic weight encoders that process weights as external data~\cite{zhou2023nft, zhou2023permutation, navon2023equivariant, lim2024learning}, we focus on enforcing structure directly on the weight space itself, through the choice of adaptation mechanism (multiplicative LoRA) and symmetry-breaking constraints (asymmetric masking). This makes the weights serve as effective representations without requiring additional encoding steps.

Generative modeling in weight space has seen significant progress, with diffusion models now capable of synthesizing functional neural networks. Weight-space generation with diffusion models \cite{wang2024neural, wang2025scaling, peebles2023learning} has been explored for generating models for image classification. Dravid \textit{et al.}~\cite{dravid2024interpreting} show that LoRA weights of diffusion models fine-tuned on human identities form an interpretable linear subspace, enabling semantic editing and sampling via PCA. With a different focus, our work explores neural field weights as representations of data that support high-quality generation and encode semantic structure.

\subsection{Implicit Neural Representation}

\camera{condensed}.
Implicit Neural Representations (INRs), or neural fields, parameterize signals as continuous functions $\Phi: \mathbb{R}^n \rightarrow \mathbb{R}^m$ via neural networks, enabling resolution-independent, modality-agnostic representations across images, shapes, and spatiotemporal data~\cite{xie2022neural, essakine2024where}. Generalizable INRs learn dataset-level priors through autodecoders~\cite{park2019deepsdf}, GANs with modulated MLP trunks~\cite{anokhin2021cips, karras2021alias, chan2020pigan, karras2019style}, and shared-layer schemes~\cite{vyas2024learning}. Since INR weights directly parameterize data, they have been explored for compression~\cite{dupont2021coin, gordon2024doh}, hypernetwork-based generation~\cite{klocek2019hypernetwork, chen2022transformers}, and meta-learned modulation representations~\cite{dupont2022functa}. In contrast to these approaches, we investigate whether independently optimized weights can directly serve as meaningful representations. Closely related to our work, HyperDiffusion~\cite{erkocc2023hyperdiffusion} trains a diffusion transformer over neural field weights to synthesize 3D and 4D shapes. We build on this line of work, investigating how the choice of adaptation mechanism and symmetry-breaking constraints affect weight space generation quality and semantic structure. A more detailed discussion is provided in the supplementary material (Section~\ref{sec:suppl_related_inr}).

\section{Method}
\label{sec:method}

We begin by establishing weight space representations via Low-Rank Adaptation of pre-trained base neural fields (Section~\ref{subsec:weight_representations}). We then introduce multiplicative LoRA, a key modification that enables effective weight space learning for neural fields (Section~\ref{subsec:multiplicative_lora}). Next, we describe the base model architecture and training procedure (Section~\ref{subsec:base_model}), followed by asymmetric masking to address permutation symmetry (Section~\ref{subsec:asymmetric_mask}). Finally, we present a hierarchical diffusion transformer for generative modeling in weight space (Section~\ref{subsec:diffusion_lora}).

\subsection{Weight Space Representation}
\label{subsec:weight_representations}

Given a dataset consisting of a collection of instances $\{\mathbf{x}_i\}_{i=1}^N$, we optimize one neural field for each instance. Each instance is then represented by the weights of its corresponding network, forming a weight space representation. 

The simplest way to represent an instance with network weights consists of using standalone MLP weights. For each instance $\mathbf{x}_i$, we fit a small MLP from scratch. The instance $\mathbf{x}_i$ is then represented with the weights of the MLP $\boldsymbol{\theta}_i$. Formally,
\begin{equation}
\boldsymbol{\theta}_i=\min_{\boldsymbol{\theta}} \mathcal{L}_{\text{recon}}
\left [ f(\mathbf{p}~|~\boldsymbol{\theta}), \mathbf{x}_i(\mathbf{p})
\right ]
\;,
\end{equation}
where $\mathbf{p}$ is a spatial coordinate.
The architecture consists of a Fourier Feature layer followed by two linear layers. We evaluate this approach as a baseline, to encourage consistency, all MLPs share the same random initialization across different instances as is done in \cite{erkocc2023hyperdiffusion}.

Encouraged by the observation that LoRA weights converge to a certain a subspace \cite{hu2022lora}, which indicates that the base network enforce a certain structure on the space of LoRA weights, we fine tune a pre-trained base model using Low-Rank Adaptation (LoRA). For each instance $\mathbf{x}_i$, we optimize LoRA parameters $\boldsymbol{\phi}_i =\{\mathbf{A}_i^l, \mathbf{B}_i^l\}_{l=1}^L$ across $L$ layers while keeping the base weights frozen. That is, we solve
\begin{equation}
\boldsymbol{\phi}_i = \min_{\boldsymbol{\phi}} \mathcal{L}_{\text{recon}}
\left [
f \left ( \mathbf{p}~|~\text{LoRA}(\mathbf{W}, \boldsymbol{\phi})
\right ), \mathbf{x}_i(\mathbf{p})
\right ] \;,
\end{equation}
where $\mathbf{W}$ denotes the frozen base model weights and $\mathcal{L}_{\text{recon}}$ is a reconstruction loss. The instance $\mathbf{x}_i$ is then represented with the LoRA weights $\boldsymbol{\phi}_i$, illustrated in Figure~\ref{fig:paradigm}. Like is done for Standalone MLP weights, all LoRA share the same initialization.

\subsection{Multiplicative LoRA}
\label{subsec:multiplicative_lora}
We employ multiplicative LoRA rather than the standard additive formulation.
Standard LoRA~\cite{hu2022lora} adapts a pre-trained weight matrix $\mathbf{W} \in \mathbb{R}^{d_{\text{out}} \times d_{\text{in}}}$ through additive low-rank updates $\mathbf{W}' = \mathbf{W} + \mathbf{B}\mathbf{A}$
where $\mathbf{A} \in \mathbb{R}^{r \times d_{\text{in}}}$ and $\mathbf{B} \in \mathbb{R}^{d_{\text{out}} \times r}$ are low-rank matrices with rank $r \ll \min(d_{\text{in}}, d_{\text{out}})$.
We introduce a multiplicative formulation that applies weight updates through elementwise multiplication as
\begin{equation}
\mathbf{W}' = \mathbf{W} \odot \mathbf{B}\mathbf{A}\;,
\end{equation}
where $\odot$ denotes elementwise multiplication. This formulation enables more effective modulation of features, analogous to successful modulation techniques in generative neural fields~\cite{anokhin2021cips, karras2021alias, chan2020pigan}. In our experiments, we demonstrate that this design is critical to obtaining good weight space structure and performance on reconstruction, generation, and discriminative tasks.

We hypothesize that the advantage of multiplicative over additive LoRA is related to \emph{feature entanglement} in neural fields. INRs synthesize signals through additive composition: linear layers combine basis functions while activation functions generate harmonics~\cite{yuce2022structured}. This additive synthesis inherently creates entangled representations. Additive LoRA exacerbates this by injecting new signal components into the already-entangled mixture, making the weight space harder to structure. In contrast, multiplicative LoRA \emph{scales existing features} rather than injecting new ones, preserving channel structure and avoiding further entanglement. This aligns with Corollary~\ref{proof:multiplicative_lora_aligned}, which shows that once permutation symmetry is eliminated, mLoRA weights are fully aligned with the base network's channel axes.

\subsection{Base Model Architecture and Training}
\label{subsec:base_model}

For LoRA-based representations, we require a strong base model that captures transferable features across the data distribution. We adopt a coordinate-based neural field architecture with multiplicative weight modulation, a design found across multiple generative neural field works~\cite{anokhin2021cips, karras2021alias, chan2020pigan}. The network consists of an MLP-based trunk, where variations across instances are injected through multiplicative weight modulation. This modulation mechanism naturally aligns with our multiplicative LoRA formulation, and the architecture is applicable, but not limited, to 2D and 3D data.

We train the base model using the variational autodecoder paradigm \cite{park2019deepsdf}. This training scheme is desirable because it requires no encoder design, aligning with the data-agnostic quality of INRs. Given a dataset $\{\mathbf{x}_i\}_{i=1}^N$, we jointly optimize the network parameters $\boldsymbol{\theta}$ and per-instance latent codes $\{\mathbf{z}_i\}_{i=1}^N$ by solving
\begin{equation}
\min_{\boldsymbol{\theta}, \{\mathbf{z}_i\}} \sum_{i=1}^N \mathcal{L}_{\text{recon}}(f_{\boldsymbol{\theta}}(\mathbf{p}, \mathbf{z}_i), \mathbf{x}_i(\mathbf{p})) + \lambda_r \|\mathbf{z}_i\|_2^2\;,
\end{equation}
where $\mathbf{p}$ represents spatial coordinates and $\lambda_r$ controls the latent code regularization. 

\begin{figure*}[t]
    \centering
    \includegraphics[width=0.9\linewidth]{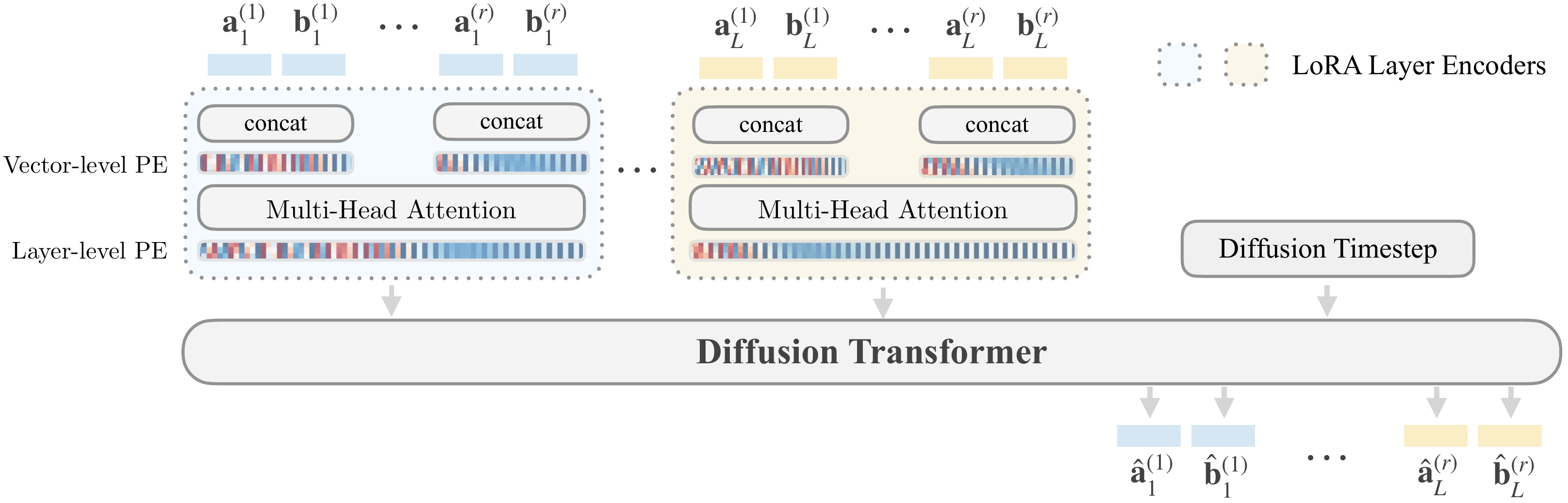}
    \caption{\textbf{Diffusion Transformer with hierarchical LoRA layer encoder architecture.} For each layer $l$, we treat vector pairs $(\mathbf{a}_l^{(i)}, \mathbf{b}_l^{(i)})$ as tokens. Vector-level positional encodings capture rank dimension indices, followed by multi-head attention that models interactions among the $r$ rank components within the layer. This hierarchical design enables the model to learn both local (within-layer) dependencies among rank components and global (cross-layer) relationships across different layers of the neural field.}
    \label{fig:transformer}
\end{figure*}

\subsection{Addressing Permutation Symmetry}
\label{subsec:asymmetric_mask}

Permutation symmetry refers to the invariance of network functions under neuron reordering, which causes functionally identical networks to occupy vastly different locations in weight space~\cite{zhao2025symmetry, gao2024revisiting, navon2023equivariant}. \camera{added} Permutation symmetry creates ambiguity, making the distribution multi-modal and difficult to learn. Removing this symmetry collapses these modes into a canonical representation, yielding a smoother, more structured weight space that could be effectively modeled.

\camera{added} Permutation symmetry has two distinct sources in our setting. \emph{External symmetries} arise from permutations of the base network neurons; these are fully eliminated by fixing a single shared base model across all instances. \emph{Internal symmetries} arise within the LoRA factors themselves: Both additive and multiplicative formulations permit permuting the $r$ rank dimensions without changing the represented function (we provide formal proofs in the supplementary material). Moreover, any invertible matrix $\mathbf{G} \in GL(r)$ gives $(\mathbf{A}\mathbf{G})(\mathbf{G}^{-1}\mathbf{B}) = \mathbf{A}\mathbf{B}$~\cite{lim2024learning}, meaning the weight space has a $GL(r)$-fold equivalence class for each represented function.

To address internal symmetry, we investigate the asymmetric masking technique of~\cite{lim2024empirical}, applied to all LoRA $\mathbf{A}$ matrices across all layers. For each layer, we randomly freeze $\sqrt{d_{\text{out}}}$ entries in each row of $\mathbf{A}$, where $d_{\text{out}}$ is the output dimension. The frozen positions are shared across all instances and training runs. For the standalone MLP and additive LoRA, frozen entries are initialized with higher variance: $\mathbf{W}_{ij} \sim \mathcal{N}(0, \kappa \mathbf{I})$, with other weights initialized with $\mathcal{N}(0, \mathbf{I})$. For multiplicative LoRA, we zero out the frozen entries: $\mathbf{A}_{ij} \leftarrow 0$, which is natural since it removes the corresponding rank component's contribution.

While asymmetric masking can be applied to all three parameterizations, it proves most effective for multiplicative LoRA in the neural field domain. For standalone MLPs and additive LoRA, the technique requires large variance $\kappa$ for the frozen entries to break symmetry effectively \cite{lim2024empirical}. However, this is problematic for neural fields, which synthesize signals through additive composition: fixing certain weights to large magnitudes creates entanglement where other weights must compensate by canceling these fixed signals, leading to difficult optimization landscapes. Multiplicative LoRA avoids this issue by zeroing out frozen entries, effectively gating off certain rank components rather than forcing compensation. This approach aligns naturally with the multiplicative structure and avoids weight entanglement. We validate this advantage empirically in our experiments.

\subsection{Diffusion Model on Weight Representations}
\label{subsec:diffusion_lora}

To evaluate the potential of the weight representations in generative tasks, we train diffusion models to learn their distribution. Following the DDPM framework, we define a forward diffusion process that gradually adds Gaussian noise to the weight representations, i.e.
\begin{equation}
q(\boldsymbol{\phi}_t | \boldsymbol{\phi}_{t-1}) = \mathcal{N}(\boldsymbol{\phi}_t; \sqrt{1-\beta_t}\boldsymbol{\phi}_{t-1}, \beta_t\mathbf{I})\;,
\end{equation}
where $\boldsymbol{\phi}$ represents the flattened weight parameters (either full MLP weights or LoRA matrices), and $\beta_t$ follows a linear schedule from $10^{-4}$ to $2 \times 10^{-2}$ over $T$ timesteps.

We parameterize the reverse process using a diffusion transformer (DiT) \cite{peebles2022scalable} that predicts the noise added to the weights. For standalone MLP weights, we adopt the architecture from \cite{erkocc2023hyperdiffusion, peebles2023learning}. For LoRA weights, we design a hierarchical LoRA layer encoder module that respects the structural properties of low-rank weight matrices, shown in Figure~\ref{fig:transformer}. Each layer $l$ is processed as follows. First, we treat each vector pair $(\mathbf{a}_l^{(i)}, \mathbf{b}_l^{(i)})$ as a token, where $\mathbf{a}_l^{(i)}$ and $\mathbf{b}_l^{(i)}$ are the $i$-th columns and rows of matrices $\mathbf{A}^{(l)}$ and $\mathbf{B}^{(l)}$, respectively. Vector-level positional encodings are then applied to capture the rank dimension index. A multi-head attention module with $r$ attention heads encodes the interactions among the $r$ vector pairs within the layer, allowing the model to learn dependencies between different rank components. Finally, layer-level positional encodings are applied to the aggregated layer representation before feeding into the main transformer.

This hierarchical design is motivated by the compositional structure of LoRA weights. Within each layer, the low-rank decomposition creates dependencies among the $r$ rank components, which the intra-layer attention module explicitly models. Different layers, however, operate at different semantic levels in the neural field, making layer-level encoding essential for capturing cross-layer relationships. This architecture naturally respects the paired nature of LoRA matrices while enabling the model to learn both local (within-layer) and global (cross-layer) weight space structure. 

With the noise prediction network $\boldsymbol{\epsilon}_{\boldsymbol{\theta}}(\boldsymbol{\phi}_t, t)$, we optimize the simplified diffusion objective
\begin{equation}
\mathcal{L} = \mathbb{E}_{t \sim \mathcal{U}(1,T), \boldsymbol{\phi}_0 \sim p_{\text{data}}, \boldsymbol{\epsilon} \sim \mathcal{N}(0,\mathbf{I})} \left[\|\boldsymbol{\epsilon} - \boldsymbol{\epsilon}_{\boldsymbol{\nu}}(\boldsymbol{\phi}_t, t)\|^2\right]\;.
\end{equation}
During inference, we use DDIM sampling to generate new weight representations, which are then used to instantiate novel neural fields.

\section{Experiments}
\label{sec:experiments}

We conduct experiments to inspect the structure of the weight space and evaluate weight space representations across reconstruction, generation, and discriminative tasks.

\noindent \textbf{Datasets.} We evaluate on two datasets. For 2D data, we use FFHQ \cite{karras2019style}, which contains high quality face images. We evaluate at the resolution of $128 \times 128$, although modest compared to current state-of-the-art in image generation, this resolution is significantly higher than what previous weight-space methods \cite{zhou2023nft, zhou2023permutation} have evaluated on and therefore more challenging. For 3D data, we use ShapeNet \cite{chang2015shapenet}, focusing on two settings: A single-category model trained on airplanes, and a multi-category model trained on ten object categories including airplanes, chairs, tables, and other common objects.

\noindent \textbf{Candidate Representations.} We compare six weight space representations: (1) \textit{MLP}: standalone MLP weights; (2) \textit{MLP-Asym}: standalone MLP weights with asymmetric masking; (3) \textit{LoRA}: additive LoRA weights; (4) \textit{LoRA-Asym}:  additive LoRA weights with asymmetric masking; (5) \textit{mLoRA}: multiplicative LoRA weights; and (6) \textit{mLoRA-Asym} multiplicative LoRA weights with asymmetric masking. This design allows us to isolate the effects of parameterization (standalone vs LoRA), operation type (additive vs multiplicative), and symmetry breaking (with or without asymmetric masks).

\begin{figure}[t]
    \centering
    \includegraphics[width=\columnwidth]{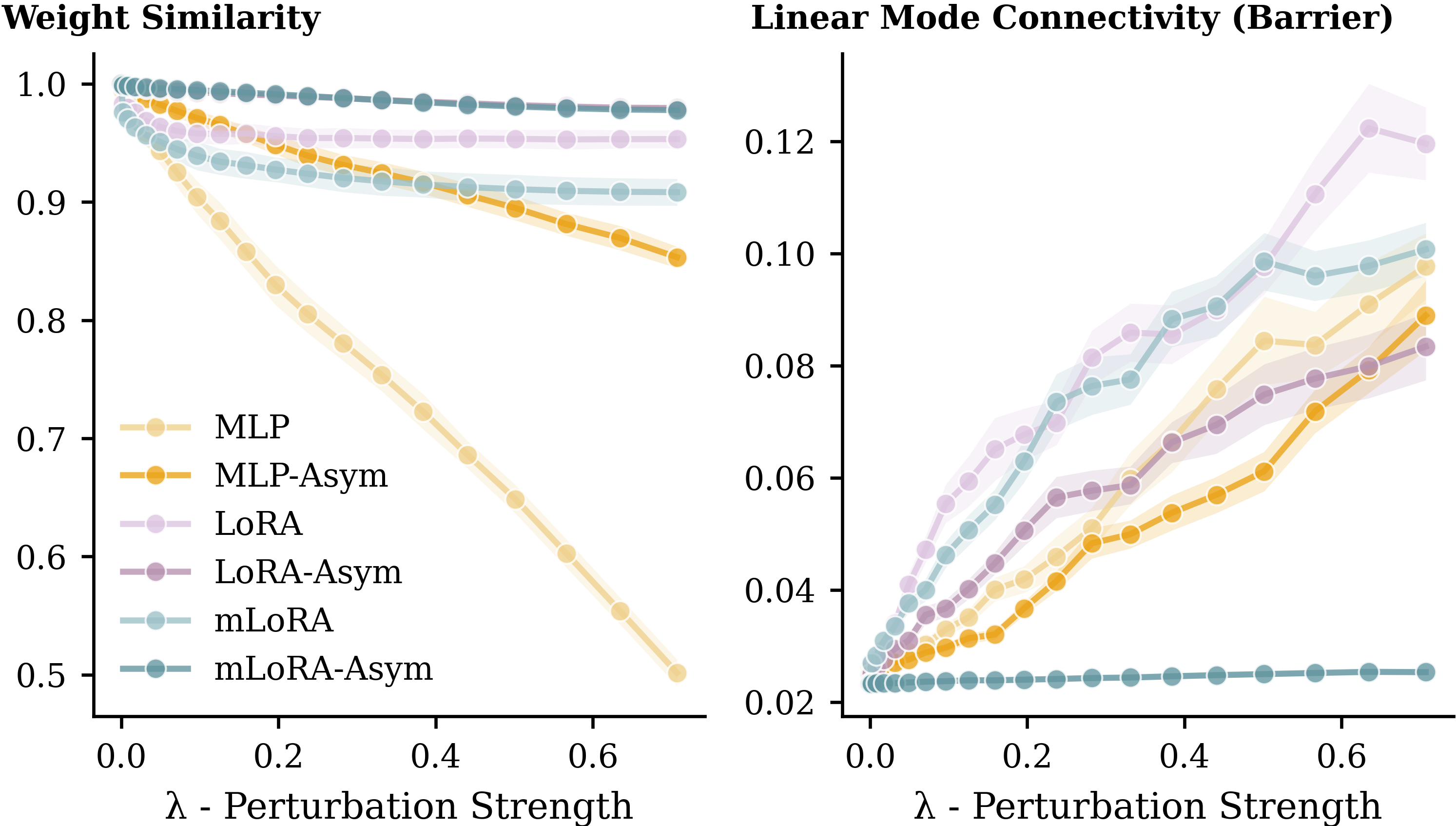}
    \caption{\textbf{Weight space structure analysis.} We measure weight similarity (cosine similarity) and the linear mode connectivity barrier (Chamfer distance) as a function of initialization perturbation strength $\lambda$. Each data point is averaged from $30$ instances, the underlying shades are indicative of standard deviation.}
    \label{fig:perturbation}
\end{figure}

\subsection{Weight Space Structure Analysis}
\label{subsec:weight_space_structure}

To understand the geometric properties of different weight spaces, we conduct a stability analysis on the ShapeNet airplane model. This experiment examines whether different random initializations lead to similar weight configurations after optimization. We independently optimize two models for each instance, starting from different initialization points. The first model is initialized with $\boldsymbol{\iota}_1 \sim \mathcal{N}(0, \mathbf{I})$, while the second is initialized with a (variance preserving) perturbed code $\sqrt{1 - \lambda^2} \boldsymbol{\iota}_1 + \lambda \boldsymbol{\iota}_2$ where $\boldsymbol{\iota}_2 \sim \mathcal{N}(0, \mathbf{I})$ and $\lambda$ controls the perturbation strength. The optimized weights from these two runs are obtained as
\begin{align*}
\boldsymbol{\phi} &= \argmin_{\boldsymbol{\phi}'} \mathcal{L}_{\text{recon}}\left [
f_{\boldsymbol{\phi}'}
\left (
\cdot |\boldsymbol{\iota}_1
\right ), \mathbf{x}
\right ]
\;, \\
\boldsymbol{\phi}_\lambda &= \argmin_{\boldsymbol{\phi}'} \mathcal{L}_{\text{recon}}\left [
f_{\boldsymbol{\phi}'}
\left (
\cdot | \sqrt{1-\lambda^2}\boldsymbol{\iota}_1 + \lambda\boldsymbol{\iota}_2
\right ),
\mathbf{x}
\right ]
\;,
\end{align*}
where $\mathbf{x}$ denotes the target instance. We evaluate two metrics to assess weight space structure. First, we measure weight similarity using the cosine similarity:
A high cosine similarity indicates that different optimization paths converge to similar weight configurations.

Second, to examine linear mode connectivity \cite{frankle2020linear}, we measure the barrier height by evaluating reconstruction quality at the midpoint of the linear interpolation path. Specifically, we compute the Chamfer Distance between the ground truth mesh vertices $\mathbf{v}_\text{gt}$ and the vertices $\mathbf{v}_\text{avg}$ extracted via marching cubes from the averaged weights, i.e.,
\begin{equation*}
b(\boldsymbol{\phi}, \boldsymbol{\phi}_\lambda) = \text{CD}\left(\mathbf{v}_\text{gt}, \mathcal{M}\left(f_{\frac{\boldsymbol{\phi}+\boldsymbol{\phi}_\lambda}{2}}\right)\right)\;,
\end{equation*}
where $\mathcal{M}(\cdot)$ denotes the marching cubes algorithm and $\text{CD}(\cdot, \cdot)$ denotes the Chamfer Distance. Figure~\ref{fig:perturbation} presents the results across varying perturbation strengths $\lambda$ for the six candidate representations.

\noindent \textbf{Results}. Figure~\ref{fig:perturbation} reveals that LoRA and mLoRA improve weight similarity compared to standalone MLPs. Specifically, weight similarity for MLP and MLP-Asym decreases approximately linearly with perturbation strength, while LoRA-based representations exhibit a saturation trend at large perturbation strengths. However, it appears that applying LoRA does not improve linear mode connectivity. This is not unexpected since permutation symmetry still exist in LoRA weights, as discussed in Section.~\ref{subsec:asymmetric_mask}. 

Although not always improving reconstruction quality, the asymmetric mask improves both weight similarity and linear mode connectivity across all parameterizations. Notably, mLoRA-Asym exhibits exceptional behavior: weight similarity remains very high and the barrier remains very low even with very different initializations. This suggests that mLoRA-Asym weights \textbf{converge to a linear mode}. The reason is likely that multiplicative LoRA weights are aligned with base networks, once permutation symmetry is eliminated, as we prove in Corollary~\ref{proof:multiplicative_lora_aligned} in the Supplementary Materials.

\camera{added discussion} Despite the application of the asymmetric mask, the use of additive LoRA weights does not exhibit as good a linear mode connectivity as mLoRA does. We hypothesize that this is related to the internal mechanisms of neural fields. Neural fields synthesize signals through iterative composition: linear layers combine basic signal components, while activation functions generate harmonics \cite{yuce2022structured}. When neural fields are optimized on individual instances, their channels become highly entangled. However, when trained on multiple instances in a generalizable setting, the network learns transferable features and exhibits greater disentanglement \cite{radford2021learning}. This observation provides additional motivation for fine-tuning a base network trained in the generative regime. Additive LoRA reintroduces entanglement by mixing features across channels, whereas multiplicative LoRA preserves the channel structure through aligned feature scaling, avoiding additional entanglement.

\subsection{Generation via Diffusion Models}

\begin{table}
\centering
\caption{Weight space generation performance on 2D FFHQ.}
\label{tab:generation_ffhq}
\resizebox{0.8\columnwidth}{!}{
\begin{tabular}{l|ccc}
\toprule
\  
~ & FD $\downarrow$ & MMD-G $\downarrow$ & MMD-P $\downarrow$  \\
\midrule
HyperDiffusion \cite{erkocc2023hyperdiffusion} 
& 0.241 & 0.158 & 1.887  \\
MLP-Asym   
& 0.287 & 0.203 & 2.423  \\
LoRA       
& 0.321 & 0.169 & 2.018  \\
LoRA-Asym  
& 0.269 & 0.157 & 1.877  \\
mLoRA      
& 0.100 & 0.056 & 0.674  \\
mLoRA-Asym 
& \textbf{0.073} & \textbf{0.039} & \textbf{0.467}  \\
\bottomrule
\end{tabular}
}
\end{table}

\begin{table*}[t]
\centering
\caption{Weight space generation performance on 3D ShapeNet. We examine a model trained on a single category \textit{Airplane} and a 10-category model \textit{Multi}. }
\label{tab:generation_shapenet}
\resizebox{\linewidth}{!}{
\begin{tabular}{l|cccccc|cccccc}
\toprule
~ & \multicolumn{6}{c|}{\textbf{ShapeNet - Airplane}} & \multicolumn{6}{c}{\textbf{ShapeNet - Multi}}           \\           
~ & mMD $\downarrow$ & COV $\uparrow$ & 1-NNA $\downarrow$ & FD $\downarrow$ & MMD-G $\downarrow$ & MMD-P $\downarrow$  & mMD $\downarrow$ & COV $\uparrow$ & 1-NNA $\downarrow$ & FD $\downarrow$ & MMD-G $\downarrow$ & MMD-P $\downarrow$  \\
\midrule
HyperDiffusion \cite{erkocc2023hyperdiffusion} 
& 2.39 & 43.6\% & 78.2\% & 0.027 & 0.009 & 0.122 & 8.64 & 41.6\% & 78.3\% & 0.117 & 0.023 & 0.219 \\
MLP-Asym   
& 2.80 & 44.8\% & 80.9\% & 0.041 & 0.018 & 0.254 & 7.77 & 46.5\% & 74.0\% & 0.085 & 0.016 & 0.157 \\
LoRA       
& 116.4 & 3.1\% & 99.9\% & 1.553 & 0.669 & 7.163 & 152.9 & 10.9\% & 99.1\% & 1.014 & 0.319 & 2.501 \\
LoRA-Asym  
& 270.6 & 2.0\% & 100\% & 1.532 & 0.823 & 7.931 & 210.0 & 1.2\% & 100\% & 1.241 & 0.437 & 2.987 \\
mLoRA      
& 1.96 & \textbf{46.2\%} & \textbf{70.5\%} & 0.049 & 0.025 & 0.359 & 5.75 & 46.4\% & 61.9\% & 0.071 & 0.011 & 0.123 \\
mLoRA-Asym 
& \textbf{1.89} & 43.4\% & 71.9\% & \textbf{0.011} & \textbf{0.003} & \textbf{0.041} & \textbf{5.52} & \textbf{49.6\%} & \textbf{58.6\%} & \textbf{0.026} & \textbf{0.004} & \textbf{0.040} \\
\bottomrule
\end{tabular}
}
\end{table*}

\begin{figure*}[t]
    \centering
    \includegraphics[width=\linewidth]{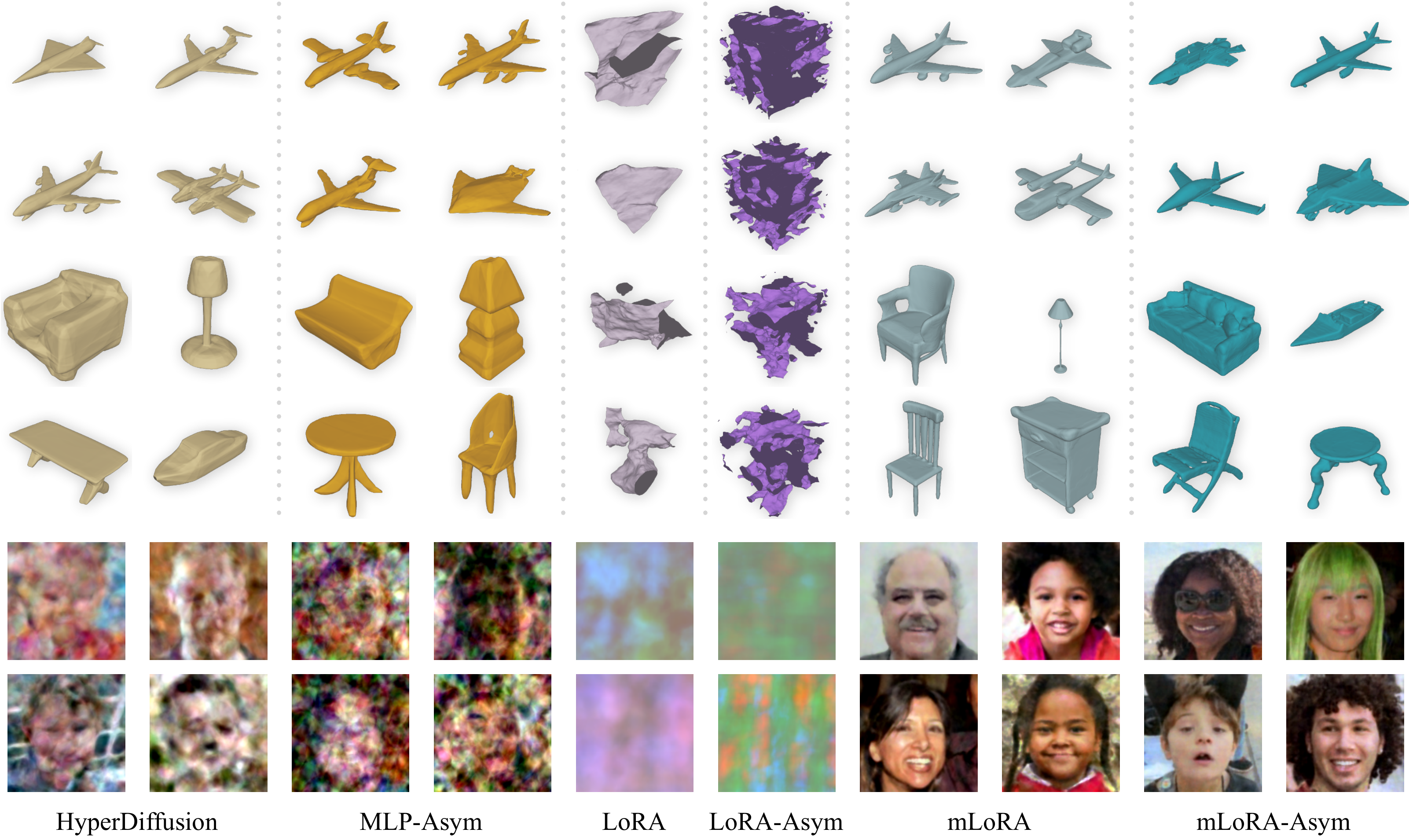}
    \caption{\textbf{Qualitative generation results.} Generated samples from diffusion models trained on different weight space representations. The top 2 rows show results generated by the \textit{Airplane} model, followed by 2 rows from the \textit{Multi}-class model. The bottom rows show 2D FFHQ generations.}
    \label{fig:generation}
\end{figure*}

We evaluate the generative capabilities of different weight space representations by training diffusion models on each parameterization. This experiment tests whether the learned weight spaces support high-quality generative modeling.

\noindent \textbf{Implementation.} For standalone MLP weights, we use the same diffusion model architecture as HyperDiffusion~\cite{erkocc2023hyperdiffusion}. For LoRA weights, we employ our hierarchical LoRA layer encoder described in Section~\ref{subsec:diffusion_lora}. For standalone MLP weights and standalone MLP weights with asymmetric masks, we adopt the initialization technique from \cite{erkocc2023hyperdiffusion}, which first fits one instance and then uses its weights to initialize all other fittings. This can be viewed as fine tuning a small MLP fitted to one instance to other instances, enabling direct comparison with the state-of-the-art weight space diffusion method.

\noindent \textbf{Evaluation Metrics.}  For both 2D and 3D data, we measure the difference between the distribution of the generated and reference samples. We compute the Fréchet distance (FD)~\cite{frechet1957distance}, as well as the Maximum Mean Discrepancy (MMD) calculated using a Gaussian RBF kernel (MMD-G) and a polynomial kernel (MMD-P), which has been shown to be more reliable and sample efficient than the Fréchet distance \cite{heusel2017gans,binkowski2018demystifying}. These metrics operate on features extracted by deep learned models, as deep learned feature extractors project the data into semantically rich embedding spaces where distances better correlate with human perception of similarity. We employ these metrics in their mathematical form rather than using modality-specific implementations like FID \cite{heusel2017gans} or KID \cite{binkowski2018demystifying}, as this allows for consistent evaluation across different data modalities. For 2D images, we use CLIP \cite{radford2021learning} as the feature extractor; for 3D shapes, we use a PointNet++ \cite{qi2017pointnet++}. For 3D shapes, following \cite{erkocc2023hyperdiffusion,vahdat2022lion, luo2021diffusion, zhou20213d}, we also report distance-based metrics: Minimum Matching Distance (mMD), Coverage (COV), and 1 Nearest Neighbor Accuracy (1-NNA). These metrics use the Chamfer Distance to measure shape similarity. Formal definitions are presented in the supplementary material.

\noindent \textbf{Results.}
Quantitative results are reported in Table~\ref{tab:generation_ffhq} for 2D FFHQ and Table~\ref{tab:generation_shapenet} for 3D ShapeNet airplane, and ShapeNet ten category datasets. Figure~\ref{fig:generation} shows visual comparisons of generated samples. The quantitative results reveal several noteworthy patterns. First, mLoRA-Asym achieves the best performance across nearly all metrics on both 2D and 3D data, capable of generating diverse samples and capturing high-frequency details. This superior generative capability correlates directly with its favorable weight space structure, suggesting that weight space geometry is crucial for effective diffusion based generation. On ShapeNet, HyperDiffusion demonstrates competent performance on the single category airplane model, but degrades substantially on the multi category setting. This suggests difficulty in modeling a diverse weight distribution spanning multiple object classes. On FFHQ, HyperDiffusion fails to produce recgnizable face images, whereas both mLoRA and mLoRA-Asym manage. This represents the first successful weight space generation for high resolution natural image generation. Previous methods have been restricted to simpler datasets such as MNIST and CIFAR. In contrast, LoRA and LoRA-Asym fail across all settings. We hypothesize that this relates to their poor weight space structure caused by entanglement of additive weights, as discuss in our weight space structure analysis. Overall, the strong correlation between weight space structure and generation performance validates that structured, well behaved weight spaces are essential for treating network parameters as effective data representations.

\subsection{Discriminative Tasks}

To evaluate the distinctiveness and semantic structure of the learned weight representations, we conduct classification and clustering experiments on the ShapeNet ten-category dataset.

\noindent \textbf{Classification.} We evaluate two classification approaches. First, we use first nearest neighbor classification, which assigns each test weight to the category of its nearest neighbor in the training set using cosine similarity. Second, we train a linear classifier using logistic regression. All classifiers take the flattened weight representations as input and predict object categories.

\noindent \textbf{Clustering.} For clustering, we apply $k$-means with $k=10$ (matching the number of categories) on the weight representations. We evaluate the clustering quality using the Adjusted Rand Index (ARI), which measures the agreement between the predicted clusters and the ground-truth categories while correcting for chance. 

\begin{table}
\centering
\caption{Classification and clustering results on the ShapeNet ten-category dataset. We report the accuracy for two classification methods and the Adjusted Rand Score for clustering. All results are statistics from 10 runs with random data split and initializations.}
\label{tab:classification}
\resizebox{\columnwidth}{!}{
\begin{tabular}{lcccc}
\toprule
& Clustering ARI $\uparrow$  & 1-NN $\uparrow$ & Logistic $\uparrow$ \\
\midrule
MLP & 39.3\% $\pm$ 3.1\% & 50.0\% $\pm$ 3.0\% & 78.1\% $\pm$ 1.3\%  \\
MLP-Asym & 48.4\% $\pm$ 3.7\% & 46.8\% $\pm$ 4.2\% & 82.1\% $\pm$ 1.2\%  \\
LoRA & 56.3\% $\pm$ 3.3\% & 75.2\% $\pm$ 1.7\% & 86.1\% $\pm$ 1.1\%  \\
LoRA-Asym & 47.4\% $\pm$ 4.4\% & 59.1\% $\pm$ 10.4\% & 84.3\% $\pm$ 0.8\%  \\
mLoRA & \textbf{67.1}\% $\pm$ 3.7\% & \textbf{85.1}\% $\pm$ 1.8\%  & \textbf{90.0}\% $\pm$ 0.7\%  \\
mLoRA-Asym & 56.5\% $\pm$ 2.7\% & 80.8\% $\pm$ 1.6\%  & 84.5\% $\pm$ 1.4\%  \\
\bottomrule
\end{tabular}
}
\end{table}

\noindent \textbf{Results.} Table~\ref{tab:classification} reports the accuracy for the two classification methods and the Adjusted Rand Score for clustering across the six candidate representations. The classification and clustering results reveal a clear progression in semantic structure across different parameterizations. LoRA outperforms standalone MLPs, mLoRA outperforms LoRA, and mLoRA delivers the best results overall, achieving $90\%$ accuracy with a linear classifier. Given that mLoRA-Asym exhibits better weight space structure in terms of linear mode connectivity, this result indicates that the discriminative power of the weight representation is not directly linked to linear mode connectivity or permutation symmetry. These results substantiate that multiplicative LoRA weight representations are capable of capturing semantic structure.

\subsection{Visualizing Weight Space Geometry}
\label{subsec:weight_space_geometry}

\begin{figure}[t]
    \centering
    \includegraphics[width=\columnwidth]{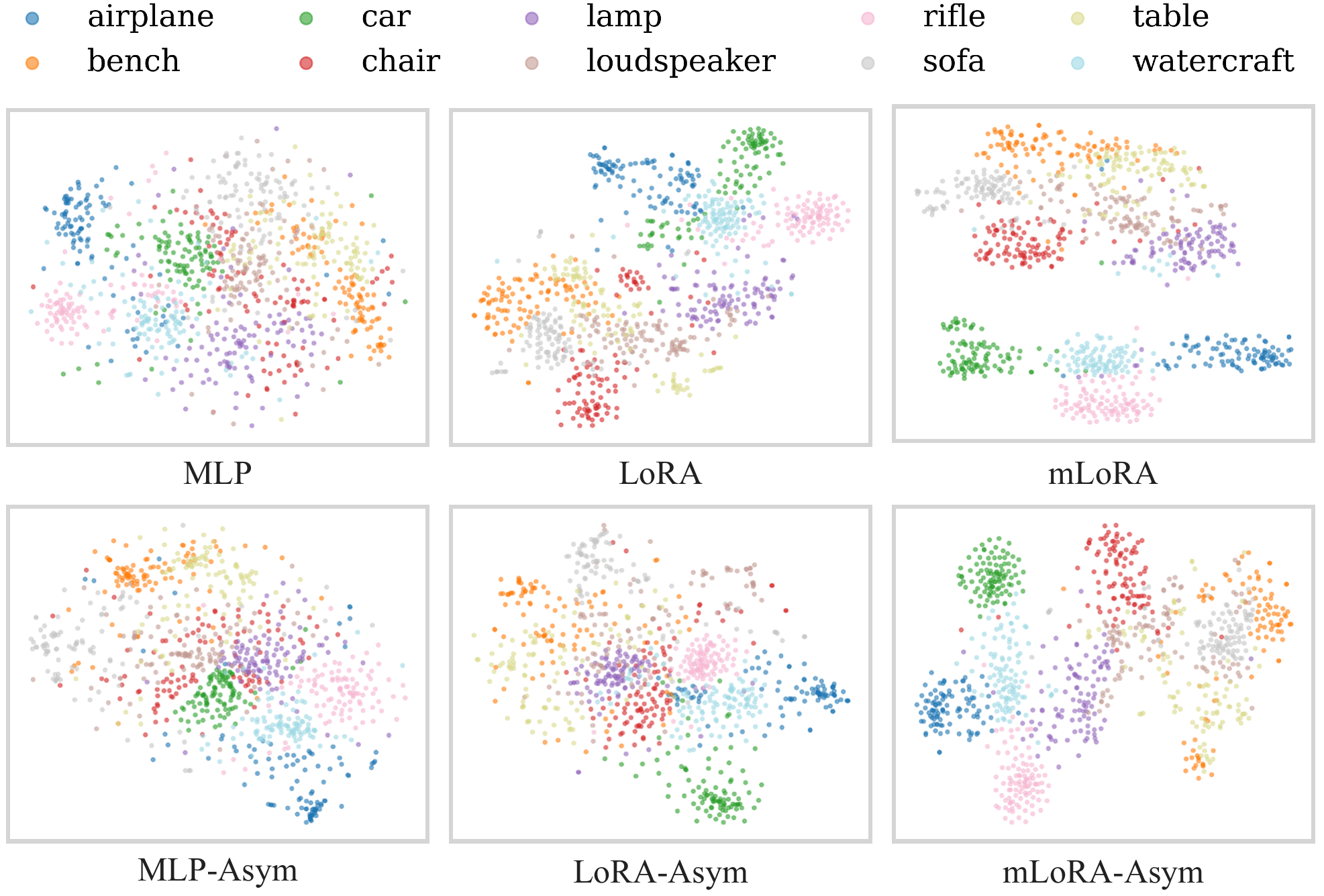}
    \caption{\textbf{t-SNE visualization of weight spaces.} Each point represents one instance from the ShapeNet ten-category dataset, colored by object category. Multiplicative LoRA weight spaces exhibit semantic structure.}
    \label{fig:tsne}
\end{figure}

We present t-SNE visualizations of the weight representations for the ten object categories in Figure~\ref{fig:tsne}. Each point encodes one instance, colored by its category. The t-SNE visualizations show that all weight representations are able to capture some level of semantic structure, as same-category instances are positioned closely. However, only multiplicative LoRA weights demonstrate clear class separation.

\begin{figure}
    \centering
    \includegraphics[width=\linewidth]{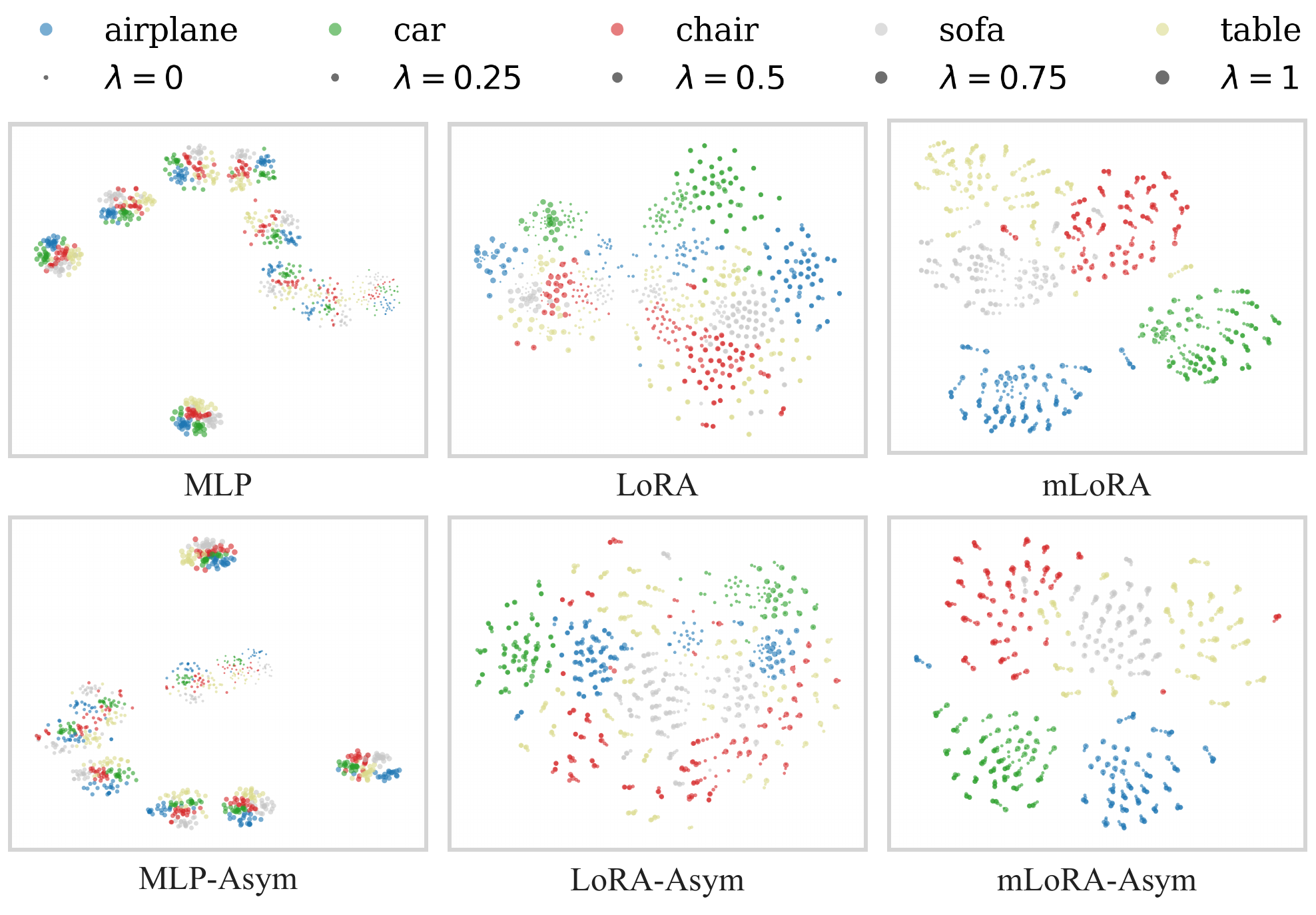}
    \caption{\textbf{t-SNE visualization of weight spaces under different initializations.} Coloring by category and sizing by the perturbation strength $\lambda$. The six representations induce different organizing hierarchies. For MLP and MLP-Asym the geometry is dominated by initialization (initialization $\rightarrow$ category $\rightarrow$ instance), whereas for mLoRA and mLoRA-Asym the semantic factor surfaces first (category $\rightarrow$ instance $\rightarrow$ initialization).}
    \label{fig:tsne_perturb}
\end{figure}

The perturbation analysis in Section~\ref{subsec:weight_space_structure} measures how a single instance's weights move as we vary its initialization. To understand how initialization, category, and instance \emph{jointly} organize the weight space, we extend this analysis to a multi-category setting. We fit $20$ instances from each of five ShapeNet categories (airplane, car, chair, sofa, and table) at $9$ perturbation strengths $\lambda$, and embed all converged weights together with t-SNE, coloring each point by its category and sizing it by $\lambda$ (Figure~\ref{fig:tsne_perturb}).

\noindent \textbf{Results.} The six representations reveal markedly different organizing hierarchies. For MLP and MLP-Asym, the weights split into nine tight modes, one per shared initialization; category appears only as sub-clusters within each mode, and individual instances sit inside those sub-clusters. The induced hierarchy is initialization $\rightarrow$ category $\rightarrow$ instance: absent any constraint to break symmetries, initialization dominates the geometry, and semantically unrelated shapes that happen to start from the same code end up close together. For mLoRA and mLoRA-Asym, this ordering is inverted. Category defines the dominant modes, and each instance occupies its own linear mode regardless of where its optimization started, so the perturbation strength $\lambda$ no longer separates the points. The hierarchy becomes category $\rightarrow$ instance $\rightarrow$ initialization, placing the semantic factor first. Additive LoRA sits in between, retaining a visible dependence on initialization. This view directly corroborates our structure analysis: breaking the weight space symmetries of multiplicative LoRA lets semantic structure surface as the primary axis of variation in weight space.

\section{Conclusion}
\label{sec:conclusion}

We have demonstrated that independently optimized neural network weights can serve as effective data representations when constrained through appropriate inductive biases. By adapting a pre-trained base model via multiplicative LoRA, we transform the chaotic parameter space into structured weight representations that exhibit semantic organization, enabling superior reconstruction quality, generation performance, and semantic structure. Remarkably, mLoRA-Asym weights converge to a linear mode during optimization, and this structured weight space geometry correlates strongly with generative performance in diffusion models. These findings challenge the view of weights as opaque byproducts of optimization and establish their viability as semantic representations for reconstruction, generation, and discriminative tasks.

{
    \small
    \bibliographystyle{ieeenat_fullname}
    \bibliography{main}

\begin{thebibliography}{47}
\providecommand{\natexlab}[1]{#1}
\providecommand{\url}[1]{\texttt{#1}}
\expandafter\ifx\csname urlstyle\endcsname\relax
  \providecommand{\doi}[1]{doi: #1}\else
  \providecommand{\doi}{doi: \begingroup \urlstyle{rm}\Url}\fi

\bibitem[Anokhin et~al.(2021)Anokhin, Demochkin, Khakhulin, Sterkin, Lempitsky,
  and Korzhenkov]{anokhin2021cips}
Ivan Anokhin, Kirill Demochkin, Taras Khakhulin, Gleb Sterkin, Victor
  Lempitsky, and Denis Korzhenkov.
\newblock Image generators with conditionally-independent pixel synthesis.
\newblock In \emph{Proceedings of the IEEE/CVF conference on computer vision
  and pattern recognition}, pages 14278--14287, 2021.

\bibitem[Bi{\'n}kowski et~al.(2018)Bi{\'n}kowski, Sutherland, Arbel, and
  Gretton]{binkowski2018demystifying}
Miko{\l}aj Bi{\'n}kowski, Danica~J Sutherland, Michael Arbel, and Arthur
  Gretton.
\newblock Demystifying mmd gans.
\newblock \emph{arXiv preprint arXiv:1801.01401}, 2018.

\bibitem[Chan et~al.(2020)Chan, Monteiro, Kellnhofer, Wu, and
  Wetzstein]{chan2020pigan}
Eric~R Chan, Marco Monteiro, Petr Kellnhofer, Jiajun Wu, and Gordon Wetzstein.
\newblock pi-gan: Periodic implicit generative adversarial networks for
  3d-aware image synthesis. arxiv e-prints, page.
\newblock \emph{arXiv preprint arXiv:2012.00926}, 2020.

\bibitem[Chang et~al.(2015)Chang, Funkhouser, Guibas, Hanrahan, Huang, Li,
  Savarese, Savva, Song, Su, et~al.]{chang2015shapenet}
Angel~X Chang, Thomas Funkhouser, Leonidas Guibas, Pat Hanrahan, Qixing Huang,
  Zimo Li, Silvio Savarese, Manolis Savva, Shuran Song, Hao Su, et~al.
\newblock Shapenet: An information-rich 3d model repository.
\newblock \emph{arXiv preprint arXiv:1512.03012}, 2015.

\bibitem[Chen and Wang(2022)]{chen2022transformers}
Yinbo Chen and Xiaolong Wang.
\newblock Transformers as meta-learners for implicit neural representations.
\newblock In \emph{European Conference on Computer Vision}, pages 170--187.
  Springer, 2022.

\bibitem[Dravid et~al.(2024)Dravid, Gandelsman, Wang, Abdal, Wetzstein, Efros,
  and Aberman]{dravid2024interpreting}
Amil Dravid, Yossi Gandelsman, Kuan-Chieh Wang, Rameen Abdal, Gordon Wetzstein,
  Alexei Efros, and Kfir Aberman.
\newblock Interpreting the weight space of customized diffusion models.
\newblock \emph{Advances in Neural Information Processing Systems},
  37:\penalty0 137334--137371, 2024.

\bibitem[Dupont et~al.(2021)Dupont, Goli{\'n}ski, Alizadeh, Teh, and
  Doucet]{dupont2021coin}
Emilien Dupont, Adam Goli{\'n}ski, Milad Alizadeh, Yee~Whye Teh, and Arnaud
  Doucet.
\newblock Coin: Compression with implicit neural representations.
\newblock \emph{arXiv preprint arXiv:2103.03123}, 2021.

\bibitem[Dupont et~al.(2022)Dupont, Kim, Eslami, Rezende, and
  Rosenbaum]{dupont2022functa}
Emilien Dupont, Hyunjik Kim, S.~M.~Ali Eslami, Danilo Rezende, and Dan
  Rosenbaum.
\newblock From data to functa: Your data point is a function and you can treat
  it like one.
\newblock In \emph{International Conference on Machine Learning}, 2022.

\bibitem[Erkoç et~al.(2023)Erkoç, Ma, Öztireli, and
  Fua]{erkocc2023hyperdiffusion}
Ziya Erkoç, Fangchang Ma, Cengiz Öztireli, and Pascal Fua.
\newblock Hyperdiffusion: Generating implicit neural fields with weight-space
  diffusion.
\newblock In \emph{International Conference on Computer Vision}, 2023.

\bibitem[Essakine et~al.(2024)Essakine, Cheng, Cheng, Zhang, Deng, Zhu,
  Sch{\"o}nlieb, and Aviles-Rivero]{essakine2024where}
Amer Essakine, Yanqi Cheng, Chun-Wun Cheng, Lipei Zhang, Zhongying Deng, Lei
  Zhu, Carola-Bibiane Sch{\"o}nlieb, and Angelica~I Aviles-Rivero.
\newblock Where do we stand with implicit neural representations? a technical
  and performance survey.
\newblock \emph{arXiv preprint arXiv:2411.03688}, 2024.

\bibitem[Frankle et~al.(2020)Frankle, Dziugaite, Roy, and
  Carbin]{frankle2020linear}
Jonathan Frankle, Gintare~Karolina Dziugaite, Daniel Roy, and Michael Carbin.
\newblock Linear mode connectivity and the lottery ticket hypothesis.
\newblock In \emph{International Conference on Machine Learning}, pages
  3259--3269. PMLR, 2020.

\bibitem[Fr{\'e}chet(1957)]{frechet1957distance}
Maurice Fr{\'e}chet.
\newblock Sur la distance de deux lois de probabilit{\'e}.
\newblock In \emph{Annales de l'ISUP}, pages 183--198, 1957.

\bibitem[Gao and Others(2024)]{gao2024revisiting}
Y Gao and Others.
\newblock Revisiting model merging: A statistical perspective.
\newblock \emph{arXiv preprint}, 2024.

\bibitem[Gordon et~al.(2024)Gordon, MacDonald, Saratchandran, and
  Lucey]{gordon2024doh}
Cameron Gordon, Lachlan~E MacDonald, Hemanth Saratchandran, and Simon Lucey.
\newblock D'oh: Decoder-only random hypernetworks for implicit neural
  representations.
\newblock In \emph{Proceedings of the Asian Conference on Computer Vision},
  pages 2507--2526, 2024.

\bibitem[Ha et~al.(2016)Ha, Dai, and Le]{ha2016hypernetworks}
David Ha, Andrew Dai, and Quoc~V Le.
\newblock Hypernetworks.
\newblock In \emph{International Conference on Learning Representations}, 2016.

\bibitem[Heusel et~al.(2017)Heusel, Ramsauer, Unterthiner, Nessler, and
  Hochreiter]{heusel2017gans}
Martin Heusel, Hubert Ramsauer, Thomas Unterthiner, Bernhard Nessler, and Sepp
  Hochreiter.
\newblock Gans trained by a two time-scale update rule converge to a local nash
  equilibrium.
\newblock \emph{Advances in neural information processing systems}, 30, 2017.

\bibitem[Hospedales et~al.(2020)Hospedales, Antoniou, Micaelli, and
  Storkey]{hospedales2020meta}
T Hospedales, A Antoniou, P Micaelli, and A Storkey.
\newblock Meta-learning in neural networks: a survey. arxiv preprint arxiv:
  200405439.
\newblock 2020.

\bibitem[Hu et~al.(2022)Hu, Shen, Wallis, Allen-Zhu, Li, Wang, Wang, Chen,
  et~al.]{hu2022lora}
Edward~J Hu, Yelong Shen, Phillip Wallis, Zeyuan Allen-Zhu, Yuanzhi Li, Shean
  Wang, Lu Wang, Weizhu Chen, et~al.
\newblock Lora: Low-rank adaptation of large language models.
\newblock \emph{ICLR}, 1\penalty0 (2):\penalty0 3, 2022.

\bibitem[Jayasumana et~al.(2024)Jayasumana, Ramalingam, Veit, Glasner,
  Chakrabarti, and Kumar]{jayasumana2024rethinking}
Sadeep Jayasumana, Srikumar Ramalingam, Andreas Veit, Daniel Glasner, Ayan
  Chakrabarti, and Sanjiv Kumar.
\newblock Rethinking fid: Towards a better evaluation metric for image
  generation.
\newblock In \emph{Proceedings of the IEEE/CVF Conference on Computer Vision
  and Pattern Recognition}, pages 9307--9315, 2024.

\bibitem[Karras et~al.(2019)Karras, Laine, and Aila]{karras2019style}
Tero Karras, Samuli Laine, and Timo Aila.
\newblock A style-based generator architecture for generative adversarial
  networks.
\newblock In \emph{Proceedings of the IEEE/CVF conference on computer vision
  and pattern recognition}, pages 4401--4410, 2019.

\bibitem[Karras et~al.(2021)Karras, Aittala, Laine, H{\"a}rk{\"o}nen, Hellsten,
  Lehtinen, and Aila]{karras2021alias}
Tero Karras, Miika Aittala, Samuli Laine, Erik H{\"a}rk{\"o}nen, Janne
  Hellsten, Jaakko Lehtinen, and Timo Aila.
\newblock Alias-free generative adversarial networks.
\newblock \emph{Advances in neural information processing systems},
  34:\penalty0 852--863, 2021.

\bibitem[Klocek et~al.(2019)Klocek, Maziarka, Wo{\l}czyk, Tabor, Nowak, and
  {\'S}mieja]{klocek2019hypernetwork}
Sylwester Klocek, {\L}ukasz Maziarka, Maciej Wo{\l}czyk, Jacek Tabor, Jakub
  Nowak, and Marek {\'S}mieja.
\newblock Hypernetwork functional image representation.
\newblock In \emph{International Conference on Artificial Neural Networks},
  pages 496--510. Springer, 2019.

\bibitem[Kofinas et~al.(2024)Kofinas, Knyazev, Zhang, Chen, Burghouts, Gavves,
  Snoek, and Zhang]{kofinas2024graph}
M Kofinas, B Knyazev, Y Zhang, Y Chen, G~J Burghouts, E Gavves, C~G~M Snoek,
  and D~W Zhang.
\newblock Graph neural networks for learning equivariant representations of
  neural networks.
\newblock In \emph{The Twelfth International Conference on Learning
  Representations}, 2024.

\bibitem[Lim et~al.(2024{\natexlab{a}})Lim, Gelberg, Jegelka, Maron,
  et~al.]{lim2024learning}
D Lim, Y Gelberg, S Jegelka, H Maron, et~al.
\newblock Learning on loras: Gl-equivariant processing of low-rank weight
  spaces for large finetuned models.
\newblock \emph{arXiv preprint arXiv:2410.04207}, 2024{\natexlab{a}}.

\bibitem[Lim et~al.(2024{\natexlab{b}})Lim, Putterman, Walters, Maron, and
  Jegelka]{lim2024empirical}
Derek Lim, Theo Putterman, Robin Walters, Haggai Maron, and Stefanie Jegelka.
\newblock The empirical impact of neural parameter symmetries, or lack thereof.
\newblock \emph{Advances in Neural Information Processing Systems},
  37:\penalty0 28322--28358, 2024{\natexlab{b}}.

\bibitem[Luo and Hu(2021)]{luo2021diffusion}
Shitong Luo and Wei Hu.
\newblock Diffusion probabilistic models for 3d point cloud generation.
\newblock In \emph{Proceedings of the IEEE/CVF conference on computer vision
  and pattern recognition}, pages 2837--2845, 2021.

\bibitem[Navon et~al.(2023)Navon, Shamsian, Achituve, Fetaya, Chechik, and
  Maron]{navon2023equivariant}
A Navon, A Shamsian, I Achituve, E Fetaya, G Chechik, and H Maron.
\newblock Equivariant architectures for learning in deep weight spaces.
\newblock In \emph{International Conference on Machine Learning}, pages
  25790--25816, 2023.

\bibitem[Ormaniec and Others(2025)]{ormaniec2025fusion}
O Ormaniec and Others.
\newblock Fusion of graph convolutional networks via optimal transport.
\newblock \emph{arXiv preprint}, 2025.

\bibitem[Park et~al.(2019)Park, Florence, Straub, Newcombe, and
  Lovegrove]{park2019deepsdf}
Jeong~Joon Park, Peter Florence, Julian Straub, Richard Newcombe, and Steven
  Lovegrove.
\newblock Deepsdf: Learning continuous signed distance functions for shape
  representation.
\newblock In \emph{Proceedings of the IEEE/CVF conference on computer vision
  and pattern recognition}, pages 165--174, 2019.

\bibitem[Peebles et~al.(2023)Peebles, Radosavovic, Brooks, Efros, and
  Malik]{peebles2023learning}
William Peebles, Ilija Radosavovic, Tim Brooks, Alexei~A Efros, and Jitendra
  Malik.
\newblock Learning to learn with generative models of neural network
  checkpoints.
\newblock \emph{arXiv preprint arXiv:2209.12892}, 2023.

\bibitem[Peebles and Xie(2022)]{peebles2022scalable}
William~S Peebles and Saining Xie.
\newblock Scalable diffusion models with transformers. 2023 ieee.
\newblock In \emph{CVF International Conference on Computer Vision (ICCV)},
  2022.

\bibitem[Qi et~al.(2017)Qi, Yi, Su, and Guibas]{qi2017pointnet++}
Charles~Ruizhongtai Qi, Li Yi, Hao Su, and Leonidas~J Guibas.
\newblock Pointnet++: Deep hierarchical feature learning on point sets in a
  metric space.
\newblock \emph{Advances in neural information processing systems}, 30, 2017.

\bibitem[Radford et~al.(2021)Radford, Kim, Hallacy, Ramesh, Goh, Agarwal,
  Sastry, Askell, Mishkin, Clark, et~al.]{radford2021learning}
Alec Radford, Jong~Wook Kim, Chris Hallacy, Aditya Ramesh, Gabriel Goh,
  Sandhini Agarwal, Girish Sastry, Amanda Askell, Pamela Mishkin, Jack Clark,
  et~al.
\newblock Learning transferable visual models from natural language
  supervision.
\newblock In \emph{International conference on machine learning}, pages
  8748--8763. PmLR, 2021.

\bibitem[Singh and Jaggi(2020)]{singh2020model}
C Singh and M Jaggi.
\newblock Model fusion via optimal transport.
\newblock In \emph{Advances in Neural Information Processing Systems}, 2020.

\bibitem[Singh et~al.()Singh, Misra, and Orvieto]{singhhyperalign}
Jaisidh Singh, Diganta Misra, and Boris Knyazev6~Antonio Orvieto.
\newblock Hyper-align: Efficient modality alignment via hypernetworks.
\newblock In \emph{Workshop on Neural Network Weights as a New Data Modality}.

\bibitem[Tancik et~al.(2020)Tancik, Srinivasan, Mildenhall, Fridovich-Keil,
  Raghavan, Singhal, Ramamoorthi, Barron, and Ng]{tancik2020fourier}
Matthew Tancik, Pratul Srinivasan, Ben Mildenhall, Sara Fridovich-Keil, Nithin
  Raghavan, Utkarsh Singhal, Ravi Ramamoorthi, Jonathan Barron, and Ren Ng.
\newblock Fourier features let networks learn high frequency functions in low
  dimensional domains.
\newblock \emph{Advances in neural information processing systems},
  33:\penalty0 7537--7547, 2020.

\bibitem[Vahdat et~al.(2022)Vahdat, Williams, Gojcic, Litany, Fidler, Kreis,
  et~al.]{vahdat2022lion}
Arash Vahdat, Francis Williams, Zan Gojcic, Or Litany, Sanja Fidler, Karsten
  Kreis, et~al.
\newblock Lion: Latent point diffusion models for 3d shape generation.
\newblock \emph{Advances in Neural Information Processing Systems},
  35:\penalty0 10021--10039, 2022.

\bibitem[Vyas et~al.(2024)Vyas, Humayun, Dashpute, Baraniuk, Veeraraghavan, and
  Balakrishnan]{vyas2024learning}
Kushal Vyas, Ahmed~I Humayun, Aniket Dashpute, Richard~G Baraniuk, Ashok
  Veeraraghavan, and Guha Balakrishnan.
\newblock Learning transferable features for implicit neural representations.
\newblock \emph{Advances in Neural Information Processing Systems},
  37:\penalty0 42268--42291, 2024.

\bibitem[Wang and Others(2025)]{wang2025scaling}
K Wang and Others.
\newblock Scaling weight space generative models.
\newblock \emph{arXiv preprint}, 2025.

\bibitem[Wang et~al.(2024)Wang, Xu, Zhou, Zang, Darrell, Liu, and
  You]{wang2024neural}
K Wang, Z Xu, Y Zhou, Z Zang, T Darrell, Z Liu, and Y You.
\newblock Neural network diffusion.
\newblock \emph{arXiv preprint arXiv:2402.13144}, 2024.

\bibitem[Xie et~al.(2022)Xie, Takikawa, Saito, Litany, Yan, Khan, Tombari,
  Tompkin, Sitzmann, and Sridhar]{xie2022neural}
Yiheng Xie, Towaki Takikawa, Shunsuke Saito, Or Litany, Shiqin Yan, Numair
  Khan, Federico Tombari, James Tompkin, Vincent Sitzmann, and Srinath Sridhar.
\newblock Neural fields in visual computing and beyond.
\newblock In \emph{Computer graphics forum}, pages 641--676. Wiley Online
  Library, 2022.

\bibitem[Yang et~al.(2024)Yang, Shen, Guo, Wang, Cao, Zhang, and
  Tao]{yang2024modelmerging}
Enneng Yang, Li Shen, Guibing Guo, Xingwei Wang, Xiaochun Cao, Jie Zhang, and
  Dacheng Tao.
\newblock Model merging in llms, mllms, and beyond: Methods, theories,
  applications and opportunities.
\newblock \emph{arXiv preprint arXiv:2408.07666}, 2024.

\bibitem[Y{\"u}ce et~al.(2022)Y{\"u}ce, Ortiz-Jim{\'e}nez, Besbinar, and
  Frossard]{yuce2022structured}
Gizem Y{\"u}ce, Guillermo Ortiz-Jim{\'e}nez, Beril Besbinar, and Pascal
  Frossard.
\newblock A structured dictionary perspective on implicit neural
  representations.
\newblock In \emph{Proceedings of the IEEE/CVF Conference on Computer Vision
  and Pattern Recognition}, pages 19228--19238, 2022.

\bibitem[Zhao et~al.(2025)Zhao, Walters, and Yu]{zhao2025symmetry}
Bo Zhao, Robin Walters, and Rose Yu.
\newblock Symmetry in neural network parameter spaces.
\newblock \emph{arXiv preprint arXiv:2506.13018}, 2025.

\bibitem[Zhou et~al.(2023{\natexlab{a}})Zhou, Yang, Burns, Cardace, Jiang,
  Sokota, Kolter, and Finn]{zhou2023permutation}
Allan Zhou, Kaien Yang, Kaylee Burns, Adriano Cardace, Yiding Jiang, Samuel
  Sokota, J~Zico Kolter, and Chelsea Finn.
\newblock Permutation equivariant neural functionals.
\newblock \emph{Advances in neural information processing systems},
  36:\penalty0 24966--24992, 2023{\natexlab{a}}.

\bibitem[Zhou et~al.(2023{\natexlab{b}})Zhou, Yang, Jiang, Burns, Xu, Sokota,
  Kolter, and Finn]{zhou2023nft}
Allan Zhou, Kaien Yang, Yiding Jiang, Kaylee Burns, Winnie Xu, Samuel Sokota,
  J~Zico Kolter, and Chelsea Finn.
\newblock Neural functional transformers.
\newblock \emph{Advances in neural information processing systems},
  36:\penalty0 77485--77502, 2023{\natexlab{b}}.

\bibitem[Zhou et~al.(2021)Zhou, Du, and Wu]{zhou20213d}
Linqi Zhou, Yilun Du, and Jiajun Wu.
\newblock 3d shape generation and completion through point-voxel diffusion.
\newblock In \emph{Proceedings of the IEEE/CVF international conference on
  computer vision}, pages 5826--5835, 2021.

\end{thebibliography}
}

\clearpage
\beginappendix

\noindent This supplementary material provides additional theoretical analysis, implementation details, and experimental results to support the main paper. Section~\ref{sec:suppl_permutation} presents formal proofs demonstrating permutation symmetry in both additive and multiplicative LoRA parameterizations.
Section~\ref{sec:suppl_related_inr} discusses related works in implicit neural representations.
Section~\ref{sec:suppl_implementation} provides comprehensive implementation details for the standalone MLP architecture, base model architecture and training, dataset preparation, and the complete training pipeline. Section~\ref{sec:suppl_metrics} formally defines the evaluation metrics used throughout our experiments. Section~\ref{sec:suppl_reconstruction} reports reconstruction quality across the weight space representations. Section~\ref{sec:suppl_ablation} presents an ablation study examining the effectiveness of the hierarchical LoRA layer encoder. Section~\ref{sec:suppl_interpolation} presents weight space interpolation experiments.
Section~\ref{sec:suppl_qualitative} provides additional qualitative generation results on FFHQ and ShapeNet datasets. Finally, Section~\ref{sec:suppl_limitation} discusses limitations and future research directions.

\section{Permutation Symmetry in LoRA}
\label{sec:suppl_permutation}

We provide a formal proof that permutation symmetry exists within both additive and multiplicative LoRA parameterizations.

\subsection{Permutation Symmetry in Additive LoRA}

\begin{theorem}
The adapted weight matrix from additive LoRA exhibits permutation symmetry with respect to the rank dimensions.
\end{theorem}

\begin{proof}
Consider the additive LoRA formulation:
\begin{equation}
\mathbf{W}' = \mathbf{W} + \mathbf{B}\mathbf{A}
\end{equation}
where $\mathbf{W} \in \mathbb{R}^{d_{\text{out}} \times d_{\text{in}}}$, $\mathbf{A} \in \mathbb{R}^{r \times d_{\text{in}}}$, and $\mathbf{B} \in \mathbb{R}^{d_{\text{out}} \times r}$.

From a neural network perspective, the operation $\mathbf{B}\mathbf{A}\mathbf{x}$ for input $\mathbf{x} \in \mathbb{R}^{d_{\text{in}}}$ can be viewed as a two layer network:
\begin{equation}
\mathbf{B}\mathbf{A}\mathbf{x} = \mathbf{B}(\mathbf{A}\mathbf{x})
\end{equation}
where $\mathbf{A}$ acts as an encoder layer compressing the input to $r$ hidden activations, and $\mathbf{B}$ acts as a decoder layer expanding back to the output dimension.

Let $\mathbf{h} = \mathbf{A}\mathbf{x} \in \mathbb{R}^r$ denote the hidden activations. Consider a permutation matrix $\mathbf{P} \in \mathbb{R}^{r \times r}$ corresponding to permutation $\pi$. We can insert $\mathbf{P}^T\mathbf{P} = \mathbf{I}$ between the two layers:
\begin{equation}
\mathbf{B}\mathbf{A}\mathbf{x} = \mathbf{B}\mathbf{P}^T\mathbf{P}\mathbf{A}\mathbf{x} = (\mathbf{B}\mathbf{P}^T)(\mathbf{P}\mathbf{A})\mathbf{x}
\end{equation}

Define $\tilde{\mathbf{A}} = \mathbf{P}\mathbf{A}$ and $\tilde{\mathbf{B}} = \mathbf{B}\mathbf{P}^T$. Then:
\begin{equation}
\mathbf{B}\mathbf{A} = \tilde{\mathbf{B}}\tilde{\mathbf{A}}
\end{equation}

This shows that $(\mathbf{B}, \mathbf{A})$ and $(\tilde{\mathbf{B}}, \tilde{\mathbf{A}})$ produce identical adapted weight matrices. Concretely, $\tilde{\mathbf{A}}$ permutes the rows of $\mathbf{A}$ (equivalently, permutes which hidden neuron each row corresponds to), and $\tilde{\mathbf{B}}$ permutes the columns of $\mathbf{B}$ by the same permutation (matching the hidden neuron reordering).

Since there are $r!$ possible permutations of $r$ hidden neurons, and each permutation produces a functionally identical network, the additive LoRA weight space exhibits $r!$-fold permutation symmetry.
\end{proof}

\begin{figure}
    \centering
    \includegraphics[width=\linewidth]{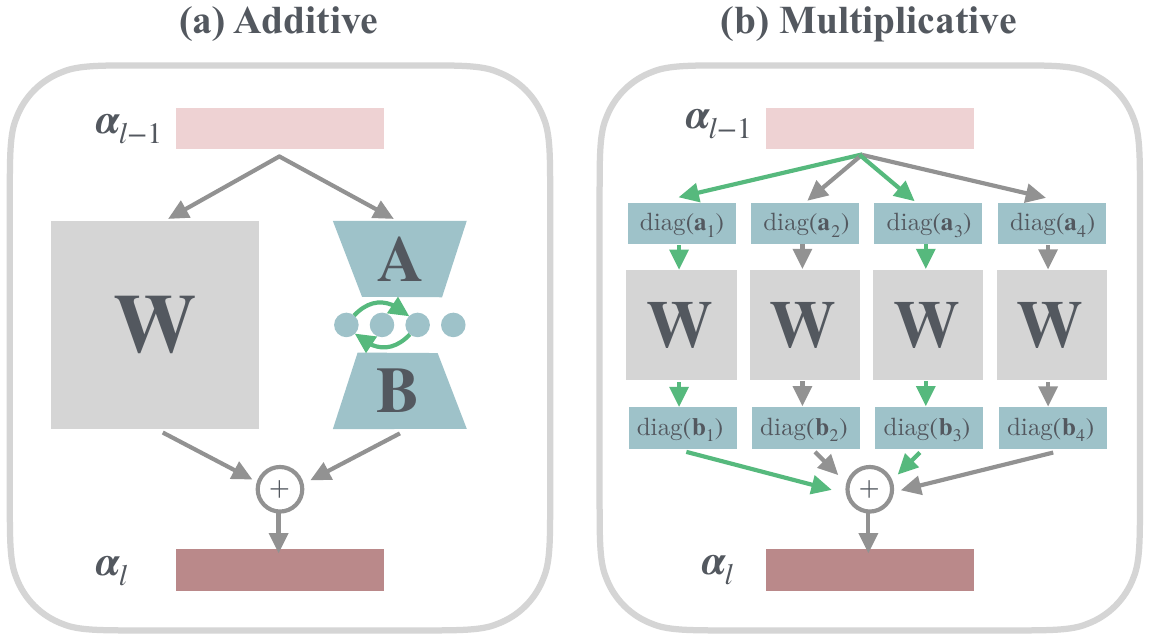}
    \caption{\textbf{Illustrating permutation symmetries within LoRA.} (a) Permutation symmetry in \textbf{additive} LoRA. Low rank matrices $\mathbf{A}$ and $\mathbf{B}$ could be seen as an encoder layer and a decoder layer. The order of the intermediate neurons could swapped without changing the output. (b) \textbf{multiplicative} LoRA could be seen as parallel pathways with different scaling factors for the input and output. The order of the pathways could be swapped without changing the output.}
    \label{fig:symmetries}
\end{figure}

\begin{remark}
This permutation symmetry is analogous to the well known permutation symmetry in standard MLPs \cite{zhou2023permutation}: reordering hidden neurons (along with their corresponding incoming and outgoing weights) does not change the function computed by the network. In LoRA, the hidden dimension is the rank $r$, and permuting this dimension induces the symmetry. Figure~\ref{fig:symmetries}(a) illustrates this symmetry, showing how the low rank matrices can be viewed as encoder and decoder layers where intermediate neurons can be reordered.
\end{remark}

\subsection{Permutation Symmetry in Multiplicative LoRA}

\begin{theorem}
The adapted weight matrix from multiplicative LoRA can be expressed as a sum of base weight matrices, each pre-multiplied and post-multiplied by diagonal matrices.
\end{theorem}

\begin{proof}
Consider the multiplicative LoRA formulation from Section~\ref{subsec:multiplicative_lora}:
\begin{equation}
\mathbf{W}' = \mathbf{W} \odot \mathbf{B}\mathbf{A}
\end{equation}
where $\mathbf{W} \in \mathbb{R}^{d_{\text{out}} \times d_{\text{in}}}$ is the base weight matrix, $\mathbf{A} \in \mathbb{R}^{r \times d_{\text{in}}}$, and $\mathbf{B} \in \mathbb{R}^{d_{\text{out}} \times r}$.

We can decompose $\mathbf{B}$ and $\mathbf{A}$ into their column and row vectors respectively:
\begin{equation}
\mathbf{B} = \begin{bmatrix}\mathbf{b}_1 & \mathbf{b}_2 & \cdots & \mathbf{b}_r\end{bmatrix}, \quad
\mathbf{A} = \begin{bmatrix}\mathbf{a}_1^T \\ \mathbf{a}_2^T \\ \vdots \\ \mathbf{a}_r^T\end{bmatrix}
\end{equation}
where $\mathbf{b}_i \in \mathbb{R}^{d_{\text{out}}}$ and $\mathbf{a}_i \in \mathbb{R}^{d_{\text{in}}}$.

Therefore:
\begin{equation}
\mathbf{W}' = \mathbf{W} \odot \left(\sum_{i=1}^r \mathbf{b}_i \mathbf{a}_i^T\right) = \sum_{i=1}^r \mathbf{W} \odot \left(\mathbf{b}_i \mathbf{a}_i^T\right)
\end{equation}

For each term in the sum, the elementwise product $\mathbf{W} \odot (\mathbf{b}_i \mathbf{a}_i^T)$ can be expressed using diagonal matrices. Let $\text{diag}(\mathbf{b}_i)$ denote the diagonal matrix with $\mathbf{b}_i$ on the diagonal, and similarly for $\text{diag}(\mathbf{a}_i)$. Then:
\begin{equation}
\mathbf{W} \odot (\mathbf{b}_i \mathbf{a}_i^T) = \text{diag}(\mathbf{b}_i) \mathbf{W} \text{diag}(\mathbf{a}_i)
\end{equation}

This can be verified by examining the $(j,k)$ entry:
\begin{align}
[\mathbf{W} \odot (\mathbf{b}_i \mathbf{a}_i^T)]_{jk} &= W_{jk} \cdot (b_i)_j \cdot (a_i)_k \\
&= (b_i)_j \cdot W_{jk} \cdot (a_i)_k \\
&= [\text{diag}(\mathbf{b}_i) \mathbf{W} \text{diag}(\mathbf{a}_i)]_{jk}
\end{align}

Therefore:
\begin{equation}
\mathbf{W}' = \sum_{i=1}^r \text{diag}(\mathbf{b}_i) \mathbf{W} \text{diag}(\mathbf{a}_i)
\label{equation:mlora_decompose}
\end{equation}

This shows that the adapted weight matrix is a sum of terms, where each term is the base weight matrix pre-multiplied and post-multiplied by diagonal matrices constructed from the LoRA parameters.
\end{proof}

\begin{corollary}
Permuting the rank indices $\{1, 2, \ldots, r\}$ with a permutation $\pi$ does not change the adapted weight matrix $\mathbf{W}'$, as summation is commutative. This implies permutation symmetry in the LoRA weight space.
\end{corollary}

This symmetry means that different configurations of LoRA parameters $\{\mathbf{a}_i, \mathbf{b}_i\}_{i=1}^r$ can represent the same function, making the weight space representation ambiguous without additional constraints. Figure~\ref{fig:symmetries}(b) visualizes this symmetry by showing how multiplicative LoRA can be interpreted as parallel pathways that can be reordered without affecting the output.

\begin{corollary}
Once permutation symmetry is eliminated, multiplicative LoRA weights are completely aligned with the channels in the base network.
\label{proof:multiplicative_lora_aligned}
\end{corollary}

\begin{proof}
In Equation~\ref{equation:mlora_decompose}, each term $\text{diag}(\mathbf{b}_i) \mathbf{W} \text{diag}(\mathbf{a}_i)$ applies channel wise modulation to the base weight matrix $\mathbf{W}$. Specifically, $\text{diag}(\mathbf{a}_i)$ scales the input channels, while $\text{diag}(\mathbf{b}_i)$ scales the output channels. This operation preserves the channel structure of $\mathbf{W}$ through element wise scaling rather than mixing features across channels. When permutation symmetry is eliminated through techniques such as asymmetric masking, each rank component $i$ is uniquely identified and cannot be arbitrarily reordered. In this regime, each pair $(\mathbf{a}_i, \mathbf{b}_i)$ corresponds to a specific modulation pattern applied to the base network channels.
\end{proof}

\begin{figure*}
    \centering
    \includegraphics[width=0.95\linewidth]{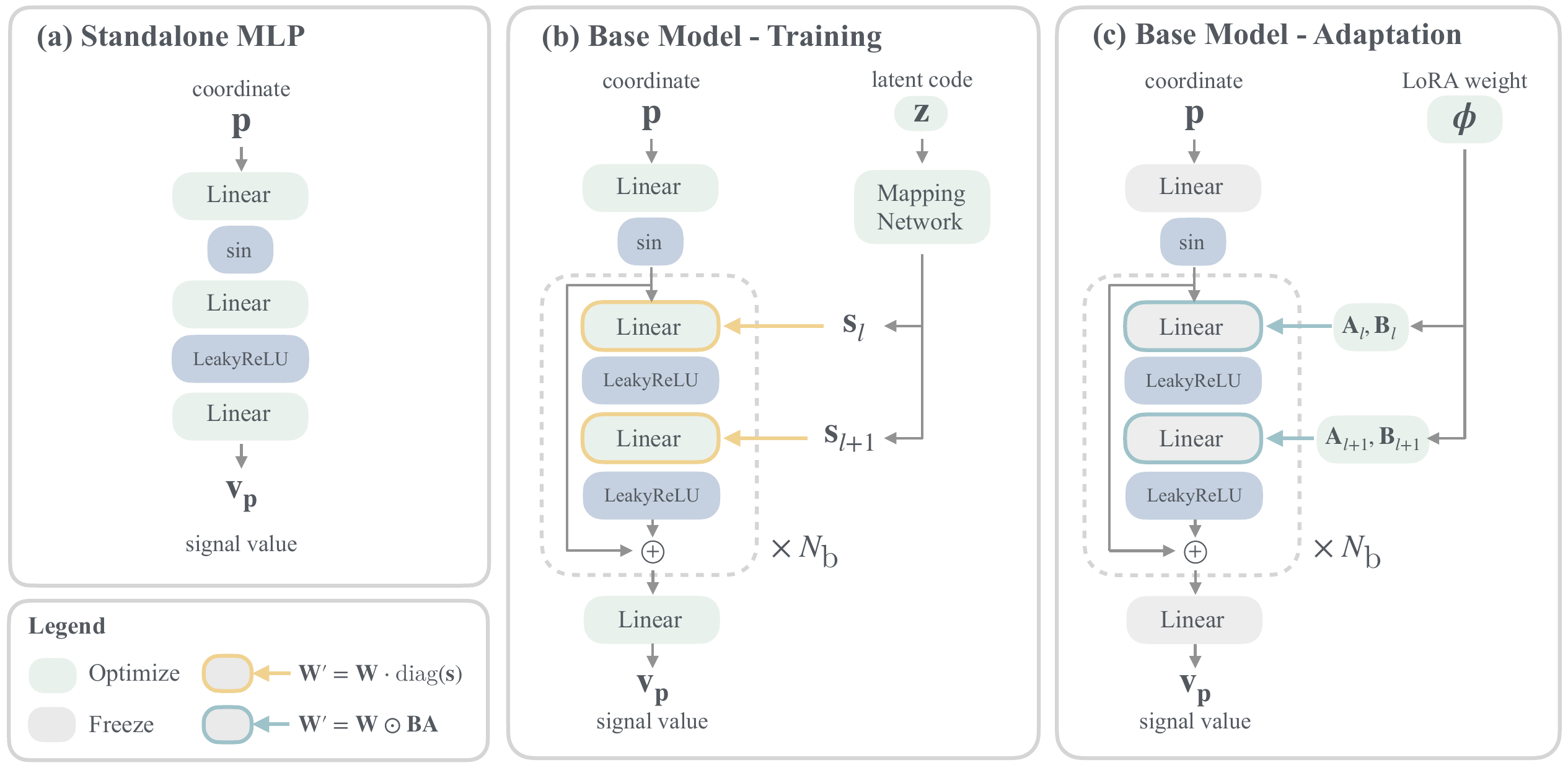}
    \caption{\textbf{Network architectures for weight space representations.} (a) Standalone MLP architecture with Fourier Feature layer followed by two linear layers. (b) Base model architecture with modulated fully connected layers. The network takes spatial coordinates $\mathbf{p}$ and style vector $\mathbf{s}$ as inputs, applying multiplicative weight modulation at each layer. (c) LoRA adaptation applied to the base model, where low rank matrices $\mathbf{A}$ and $\mathbf{B}$ adapt the frozen base weights.}
    \label{fig:nn_arch}
\end{figure*}

\section{Related Works - {\normalsize Implicit Neural Representations}}
\label{sec:suppl_related_inr}

Implicit Neural Representations (INRs), also known as neural fields, are continuous functions parameterized by neural networks that map coordinates to signal values. INRs represent signals as continuous functions $\Phi: \mathbb{R}^n \rightarrow \mathbb{R}^m$, where a neural network maps $n$-dimensional coordinates to $m$-dimensional quantities. This paradigm enables resolution-independent and modality agnostic representations of complex signals \cite{xie2022neural, essakine2024where}, employing identical network architectures across diverse modalities including 1D audio, 2D images, 3D shapes, and even 4D spatiotemporal data.

Beyond single-instance fitting, generalizable INRs have been proposed to learn priors across datasets through approaches based on autoencoders \cite{park2019deepsdf}, generative adversarial networks (GANs) \cite{anokhin2021cips, karras2021alias, chan2020pigan} and shared layers \cite{vyas2024learning}. The GAN-based works could be seen as extensions of the StyleGAN \cite{karras2019style} paradigm into the realm of neural fields. They generate different instances by modulating an MLP trunk, which we use as the base network for fine-tuning.

Because INRs parameterize data as neural network functions, the weights offer a direct pathway to data representation. This perspective has practical applications in compression, with methods \cite{dupont2021coin, gordon2024doh} demonstrating competitive compression ratios by storing quantized network parameters instead of raw data. However, whether the collection of weights could encode semantic structure remains an open question. Another line of work employs hypernetworks \cite{klocek2019hypernetwork} and transformers \cite{chen2022transformers} to predict INR weights from input data via learned mappings as a way of data generation. Dupont \textit{et al.} \cite{dupont2022functa} propose \textit{functa}, which meta-learns a shared SIREN base network and represents each data point as a low-dimensional shift modulation vector for downstream tasks including generation and classification.

In contrast to all the above approaches, we investigate whether independently optimized weights can directly serve as meaningful representations. HyperDiffusion \cite{erkocc2023hyperdiffusion} trains a diffusion transformer to generate neural field weights as a means of synthesizing 3D shapes and 4D animated shapes. Our work builds on this to inspect factors that affect weight space generation performance and to explore semantic structures in neural field weights.

\section{Implementation Details}
\label{sec:suppl_implementation}

\subsection{Standalone MLP}

As shown in Figure~\ref{fig:nn_arch}(a), The standalone MLP is a Fourier Feature \cite{tancik2020fourier} layer $\boldsymbol{\alpha}_1 = \sin(\omega_0 \cdot (\mathbf{W}_1 \mathbf{p} + \mathbf{b}_1))$ followed by $2$ linear layers. \begin{itemize}
    \item $\omega_0$ is the frequency scaling factor for the Fourier Feature layer. For 2D FFHQ, we set $\omega_0=32$; For 3D ShapeNet, we set $\omega_0=1$. This value is chosen empiracally and shared with its LoRA-based counterparts.
    \item $N_\text{hidden}$ is the number of hidden features in the linear layers. For the 2D FFHQ model, we use $N_\text{hidden}=94$ for all layers; For the 3D ShapeNet, we use $N_\text{hidden}=99$ for all layers, in order to have approximately the same number of learnable parameters as their LoRA-based counterparts.
\end{itemize} 

As noted by Erkoç \textit{et al.} \cite{erkocc2023hyperdiffusion}, an initialization trick is employed to ensure diffusion model generalization. Specifically, an MLP with weights $\boldsymbol{\phi}_0$ is fitted to one instance from the dataset, and used as a shared initialization for all the rest of the fittings $\boldsymbol{\iota}_i = \boldsymbol{\phi}_0$. In other words, other instances are fitted by fine-tuning the weights from the first instance. This trick is crucial to the performance of HyperDiffusion, and therefore we employ it in the FFHQ and ShapeNet airplane experiments. However, in the multi-category experiment, it does not make sense to initialize a \textit{chair} fitting with weights from an \textit{airplane}, therefore we do not use this trick and instead use a shared random initilization for all the fittings, like is done for the LoRA-based weight representations.

For each instance, we run $10$k steps, each step with $8,192$ points with an Adam optimizer, where learning rate adaptively decay from $10^{-2}$ to $10^{-5}$. The same optimizer and hyperparameters are used for the LoRA-based weight representations.

\subsection{Base Model Architecture}

The base model follows the modulated neural field architecture widely adopted in generative neural field works \cite{anokhin2021cips, karras2021alias, chan2020pigan}, as illustrated in Figure~\ref{fig:nn_arch}(b). The architecture consists of a mapping network and a synthesis network. The mapping network is a multilayer perceptron that transforms a latent code $\mathbf{z} \in \mathbb{R}^{d_z}$ into intermediate style vectors $\{\mathbf{s}_l\}_{l=1}^L$, where $L$ is the number of synthesis layers. For 2D FFHQ, we use $d_z=256$ and an 8 layer mapping network with hidden dimension 256. For 3D ShapeNet, we use $d_z=128$ and a 4 layer mapping network with hidden dimension 256.

The synthesis network takes spatial coordinates $\mathbf{p}$ as input and produces signal values through a sequence of synthesis blocks, each block contains 2 modulated fully connected. The blocks are connected with residual connections to ensure gradient flow. Each layer $l$ first applies weight modulation, where the weight matrix $\mathbf{W}_l$ is scaled by a learned affine transformation of the style vector: $\mathbf{W}^\prime_l = \mathbf{W}_l \cdot \text{diag}(\mathbf{s}_l)$, where $\mathbf{s}_l = \mathbf{A}_l \mathbf{s} + \mathbf{b}_l$ is computed from the style vector via an affine transformation. The modulated weights are then normalized per output channel as $w^\prime_{ijk} = w^\prime_{ijk} / \sqrt{\sum_{i,k} (w^\prime_{ijk})^2 + \epsilon}$, where $\epsilon = 10^{-8}$ for numerical stability. The layer then computes activations as $\boldsymbol{\alpha}_{l+1} = \text{ReLU}(\mathbf{W}^\prime_l \boldsymbol{\alpha}_l + \mathbf{b}_l)$. For 2D FFHQ, we use 6 synthesis blocks with channel dimensions $256$. For 3D ShapeNet, we use 4 synthesis blocks with channel dimensions $512$.

\subsection{Base Model Training}

We devise a multistage progressive training strategy that gradually increases sampling resolution while decreasing batch size. Early stages use large batch sizes with low resolution (batch size of $256$ with $2,048$ points per instance) to establish the latent code manifold, while late stages use small batch sizes with high resolution (batch size of $16$ with $32,768$ points per instance) to capture fine details. This strategy ensures stable latent code initialization while maintaining computational efficiency.

We train $350$k steps with an Adam optimizer, where the learning rate gradually decay from $10^{-3}$ to $10^{-5}$ in 5 stages. For regularization factor we use $\lambda_r=10^{-4}$ Exponential moving average is applied on the base model weights.

\subsection{Dataset}

For FFHQ, we use the first $5,000$ samples from the dataset for all our experiments. For ShapeNet airplane, we use all $4,045$ samples. To create the ShapeNet 10-category dataset, we select the top 10 categories with the most samples and then randomly sample $500$ instances from each category.

\subsection{Pipeline}

Algorithm~\ref{alg:weight_diffusion} describes the complete pipeline for weight space representation learning and generation. The process consists of three stages: First, we train a base model using variational autodecoding for LoRA based representations. Second, we construct a dataset of weight representations by fitting neural fields to individual instances. For standalone MLP, an initialization trick is employed where one instance is first fitted and then used to initialize all other fittings. For LoRA based representations, all instances share the same random initialization. Third, we train a diffusion model on the collected weight representations to enable generation of novel instances.

\begin{algorithm}[t]
\caption{Pipeline for weight space learning and generation.}
\label{alg:weight_diffusion}
\begin{algorithmic}[1]
\STATE \textbf{Input:} Dataset $\{\mathbf{x}_i\}_{i=1}^N$, parameterization type $\tau \in \{\text{MLP}, \text{LoRA}, \text{mLoRA}\}$
\STATE \textbf{Output:} Trained diffusion model $\boldsymbol{\epsilon}_{\boldsymbol{\theta}}$

\vspace{0.5em}
\STATE \textbf{Stage 1: Base Model Training} (for LoRA/mLoRA only)
\IF{$\tau \in \{\text{LoRA}, \text{mLoRA}\}$}
    \STATE Initialize base model weights $\boldsymbol{\theta}$ and latent codes $\{\mathbf{z}_i\}_{i=1}^N$
    \STATE Jointly optimize $\boldsymbol{\theta}$ and $\{\mathbf{z}_i\}$ via variational autodecoding:
    \STATE$\boldsymbol{\theta}^*, \{\mathbf{z}_i^*\} \leftarrow \argmin$ \\
    \quad $\sum_{i=1}^N \mathcal{L}_{\text{recon}}(f(\mathbf{p}, \mathbf{z}_i | \boldsymbol{\theta}), \mathbf{x}_i(\mathbf{p})) + \lambda_r \|\mathbf{z}_i\|_2^2$
    \STATE Freeze base model: $\boldsymbol{\theta} \leftarrow \boldsymbol{\theta}^*$
\ENDIF

\vspace{0.5em}
\STATE \textbf{Stage 2: Instance Fitting}
\IF{$\tau = \text{MLP}$}
    \STATE Fit one instance: \\
    $\boldsymbol{\phi}_0 \leftarrow \argmin_{\boldsymbol{\phi}} \mathcal{L}_{\text{recon}}(f(\mathbf{p}~|~\boldsymbol{\phi}), \mathbf{x}_1(\mathbf{p}))$
    \FOR{$i=2$ to $N$}
        \STATE Initialize: $\boldsymbol{\phi}_i \leftarrow \boldsymbol{\phi}_0$
        \STATE Optimize: \\
        $\boldsymbol{\phi}_i \leftarrow \argmin_{\boldsymbol{\phi}} \mathcal{L}_{\text{recon}}(f(\mathbf{p}~|~\boldsymbol{\phi}), \mathbf{x}_i(\mathbf{p}))$
    \ENDFOR
\ELSE
    \STATE Sample shared random initialization: $\boldsymbol{\iota}_0 \sim \mathcal{N}(0, \mathbf{I})$
    \FOR{$i=1$ to $N$}
        \STATE Initialize: $\boldsymbol{\phi}_i \leftarrow \boldsymbol{\iota}_0$
        \STATE Optimize: \\
        $\boldsymbol{\phi}_i \leftarrow \argmin_{\boldsymbol{\phi}} \mathcal{L}_{\text{recon}}(f(\mathbf{p}~|~\text{LoRA}(\mathbf{W}, \boldsymbol{\phi})), \mathbf{x}_i(\mathbf{p}))$
    \ENDFOR
\ENDIF

\vspace{0.5em}
\STATE \textbf{Stage 3: Diffusion Model Training}
\STATE Initialize diffusion model parameters $\boldsymbol{\nu}$
\WHILE{not converged}
    \STATE Sample $t \sim \mathcal{U}(1,T)$, $\boldsymbol{\phi}_0 \sim \{\boldsymbol{\phi}_i\}_{i=1}^N$, $\boldsymbol{\epsilon} \sim \mathcal{N}(0,\mathbf{I})$
    \STATE Compute $\boldsymbol{\phi}_t = \sqrt{\bar{\alpha}_t}\boldsymbol{\phi}_0 + \sqrt{1-\bar{\alpha}_t}\boldsymbol{\epsilon}$
    \STATE Update: $\boldsymbol{\nu} \leftarrow \boldsymbol{\nu} - \nabla_{\boldsymbol{\nu}} \|\boldsymbol{\epsilon} - \boldsymbol{\epsilon}_{\boldsymbol{\nu}}(\boldsymbol{\phi}_t, t)\|^2$
\ENDWHILE
\STATE \textbf{return} $\boldsymbol{\epsilon}_{\boldsymbol{\theta}}$
\end{algorithmic}
\end{algorithm}

For the diffusion model architecture, we use a standard Diffusion Transformer (DiT) \cite{peebles2022scalable} for standalone MLP weights. For LoRA weights, we employ the hierarchical LoRA layer encoder described in Section~\ref{subsec:diffusion_lora} of the main paper. During training, we use the DDPM objective with a linear noise schedule from $\beta_1 = 10^{-4}$ to $\beta_T = 2 \times 10^{-2}$ over $T = 500$ timesteps. For generation, we use DDIM sampling with 100 steps to efficiently sample new weight representations. The diffusion transformer has 2880 hidden size (i.e., the size of
each token after linear projection or layer encoder), 12 layers, and 16 self-attention heads. We train the diffusion transformer with batch size of 256 and learning rate of $2 \times 10^{-4}$ for $6000$ epochs until convergence.

\section{Evaluation Metrics for Generation}
\label{sec:suppl_metrics}

We calculate distributional metrics for both 2D and 3D. For 3D, we also calculate distance-based metrics. All metrics are calculated between $2,048$ generated samples and $2,048$ reference samples.

\subsection{Distributional Difference}

These metrics operate on features extracted by deep learned models, as deep learned feature extractors project the data into semantically rich embedding spaces where distances better correlate with human perception of similarity. We employ these metrics in their mathematical form rather than using modality-specific implementations like FID \cite{heusel2017gans} or KID \cite{binkowski2018demystifying}, as this allows for consistent evaluation across different data modalities. For 2D images, we use CLIP \cite{radford2021learning} as the feature extractor; for 3D shapes, we use a PointNet++ \cite{qi2017pointnet++}.

Given generated distribution $P$ and reference distribution $Q$, to make the metric more comparable, we first normalize the extracted features by
\begin{align*}
\boldsymbol{\rho} \leftarrow \frac{\boldsymbol{\rho} - \mu_Q}{\sigma_Q}
\end{align*}
for both the generated set and reference set, where $\mu_Q$ and $\sigma_Q$ are the scalar mean and standard deviation calculated from the reference set.

\noindent \textbf{Fréchet Distance (FD)} \cite{frechet1957distance} measures the distance between two multivariate Gaussian distributions fitted to feature representations of generated and reference samples. Given feature representations from a pretrained network, we compute the mean $\boldsymbol{\mu}_P$ and covariance $\boldsymbol{\Sigma}_P$ for generated distribution $P$ and mean $\boldsymbol{\mu}_Q$ and covariance $\boldsymbol{\Sigma}_Q$ for reference distribution $Q$. The Fréchet Distance is then computed as:
\begin{align*}
\text{FD}(P, Q) = \frac{1}{N_\text{feature}} 
[ 
& \|\boldsymbol{\mu}_P - \boldsymbol{\mu}_Q\|_2^2 + \\
& \text{Tr}\left(\boldsymbol{\Sigma}_P + \boldsymbol{\Sigma}_Q - 2(\boldsymbol{\Sigma}_P \boldsymbol{\Sigma}_Q)^{1/2}\right) 
] \;.
\end{align*}
where $N_\text{feature}$ is the feature dimension of the feature extractor.

\noindent \textbf{Maximum Mean Discrepancy (MMD)} with respect to a positive definite kernel $\psi$ is defined by:
\begin{align*}
    \text{MMD}(P, Q) = \mathbb{E}_{\mathbf{x}, \mathbf{x}'}[\psi(\mathbf{x}, \mathbf{x}')] &+ \mathbb{E}_{\mathbf{y}, \mathbf{y}'}[\psi(\mathbf{y}, \mathbf{y}')] \nonumber  \\
    &- 2 \mathbb{E}_{\mathbf{x}, \mathbf{y}}[\psi(\mathbf{x}, \mathbf{y})],
\end{align*}
where $\mathbf{x}, \mathbf{x}' \sim P$ are samples from the generated distribution and $\mathbf{y}, \mathbf{y}' \sim Q$ are samples from the reference distribution. This metric does not make the multivariate Gaussian assumption and is reported to be more reliable and sample efficient \cite{jayasumana2024rethinking, binkowski2018demystifying}. We calculate this metric with 2 types of kernels.
\begin{enumerate}
    \item Polynomial kernel (MMD-P) \\
    $\psi_p(\mathbf{x}, \mathbf{y}) = (\gamma_p \cdot \mathbf{x}^T \mathbf{y} + c)^d$ with degree $d=3$ and offset $c=1$. Following the practice of KID \cite{binkowski2018demystifying}, we choose $\gamma_p=1/N_\text{feature}$.
    \item Gaussian RBF kernel (MMD-G) \\
    $\psi_g(\mathbf{x}, \mathbf{y}) = \exp(-\gamma_g \|\mathbf{x} - \mathbf{y}\|_2^2)$ with $\gamma_g = 1 / (2\sigma_g^2)$; we choose $\sigma_g=N_\text{feature}$.
\end{enumerate}

\subsection{Distance-based Metrics for 3D Shapes}

For 3D shapes, we calculate distance-based metrics following \cite{erkocc2023hyperdiffusion,vahdat2022lion, luo2021diffusion, zhou20213d}. We denote the distance function as $D(\mathbf{x}, \mathbf{y})$ for the Chamfer Distance between two shapes $\mathbf{x}$ and $\mathbf{y}$. The metrics are defined as:
\begin{align*}
\text{mMD}(P, Q) &= \mathbb{E}_{\mathbf{y} \sim Q} \left[ \min_{\mathbf{x} \sim P} D(\mathbf{x}, \mathbf{y}) \right], \\
\text{COV}(P, Q) &= \frac{\vert \{ \argmin_{\mathbf{y} \sim Q} D(\mathbf{x}, \mathbf{y}) \vert \mathbf{x} \sim P \} \vert}{\vert Q \vert}, \\
\text{1-NNA}(P, Q) &= \frac{\sum_{\mathbf{x} \sim P} \mathbbm{1}[\mathbf{N}_{\mathbf{x}} \sim P] + \sum_{\mathbf{y} \sim Q} \mathbbm{1}[\mathbf{N}_{\mathbf{y}} \sim Q] }{\vert P \vert + \vert Q \vert},
\end{align*}
where in the 1-NNA metric $\mathbf{N}_{\mathbf{x}}$ is the shape that is closest to $\mathbf{x}$ in both generated and reference distributions, that is,
$$\mathbf{N}_{\mathbf{x}} = \argmin_{\mathbf{z} \sim P \cup Q} D(\mathbf{x}, \mathbf{z}).$$
For mMD, lower is better; for COV, higher is better; for 1-NNA, 50\% is optimal.

\section{Reconstruction Experiments}
\label{sec:suppl_reconstruction}

\begin{table}[t]
\centering
\caption{Reconstruction quality. We report the PSNR for 2D FFHQ, the Chamfer Distance $\times 10^{-2}$ for 3D ShapeNet, and the number of trainable parameters in each weight space representation. Parameters frozen by the asymmetric mask are reduced from the count.}
\label{tab:reconstruction}
\resizebox{\columnwidth}{!}{
\begin{tabular}{l|cc|ccc}
\toprule
           & \multicolumn{2}{c|}{\textbf{FFHQ}} & \multicolumn{3}{c}{\textbf{ShapeNet}}            \\

           & PSNR $\uparrow$     & \# Params    & CD-A $\downarrow$ & CD-M $\downarrow$ & \# Params \\
\midrule
MLP        & 35.11 & 27,357 & 2.57 & 3.78 & 30,196   \\
MLP-Asym   & 33.28 & 24,537 & 2.64 & 4.00 & 27,226      \\
LoRA       & 35.69 & 27,395 & 2.44 & 3.39 & 29,696         \\
LoRA-Asym  & 24.63 & 26,307 & 2.46 & 3.44 & 27,539      \\
mLoRA      & 35.65 & 27,395 & 2.45 & 3.49 & 29,696      \\
mLoRA-Asym & \textbf{36.91} & 26,307 & \textbf{2.41} & \textbf{3.35} & 27,539     \\
\bottomrule
\end{tabular}
}
\vspace{-0.5em}
\end{table}

The reconstruction task directly corresponds to the fitting procedure described in Section~\ref{subsec:weight_representations}, where each weight space representation is optimized to reconstruct individual instances. For 2D images, we measure the PSNR (higher is better). For 3D shapes, we measure the Chamfer Distance (lower is better). We also report the number of learnable parameters for each representation.

\noindent \textbf{Results}. As shown in Table~\ref{tab:reconstruction}, LoRA, mLoRA and mLoRA-Asym achieve better reconstruction quality while maintaining a compact parameter count, compared to the MLP-based weight representations. We attribute this to the inductive bias provided by the base network, which captures transferable features across instances, and fine tuning effectively leverages these shared representations with minimal adaptation parameters.
Multiplicative LoRA perform better than its additive counterpart, echoing the wide application of multiplicative modulation in generative neural fields \cite{anokhin2021cips, chan2020pigan, karras2021alias}.
Surprisingly, mLoRA-Asym outperforms mLoRA despite having certain parameters frozen by the asymmetric mask. Our hypothesis is that the masking further reduces parameter entanglement and thereby improves the reconstruction accuracy. Also notably, LoRA-Asym performs poorly on FFHQ, likely due to increased parameter entanglement caused by large variance $\kappa=6$ (empirically determined) of the frozen weights, as discussed in Section ~\ref{subsec:asymmetric_mask}.

\section{Ablation Study}
\label{sec:suppl_ablation}

We examine the effectiveness of the hierarchical LoRA layer encoder introduced in Section~\ref{subsec:diffusion_lora}. To isolate its contribution, we train a baseline diffusion model without the layer encoder on the ShapeNet multi-category dataset. This baseline treats each weight matrix as an independent token, directly feeding flattened LoRA matrices into the transformer without the hierarchical processing that models within layer rank dependencies and cross layer relationships.

Table~\ref{tab:generation_ablation} compares the baseline against our full model with the hierarchical layer encoder. The results demonstrate that the layer encoder provides substantial improvements across all metrics. The full model achieves higher coverage (49.6\% vs 47.7\%), better 1-NNA (58.6\% vs 61.2\%), and significantly better distributional metrics: FD improves from 0.049 to 0.026, MMD-G from 0.008 to 0.004, and MMD-P from 0.098 to 0.040. These improvements confirm that explicitly modeling the compositional structure of LoRA weights, where rank components within each layer interact and different layers encode different semantic levels, is essential for effective weight space generation. The hierarchical design enables the diffusion model to respect the intrinsic organization of neural field weights, leading to higher quality generated samples.

\begin{table}[t]
\centering
\caption{Ablation study of LoRA Layer Encoder on 3D ShapeNet - 10 category. }
\label{tab:generation_ablation}
\resizebox{\linewidth}{!}{
\begin{tabular}{l|cccccc}
\toprule
~ & \multicolumn{6}{c}{\textbf{ShapeNet - Multi}}           \\           
~ & mMD $\downarrow$ & COV $\uparrow$ & 1-NNA $\downarrow$ & FD $\downarrow$ & MMD-G $\downarrow$ & MMD-P $\downarrow$ \\
\midrule
w/o 
& \textbf{5.18} & 47.7\% & 61.2\% & 0.049 & 0.008 & 0.098 \\
with 
& 5.52 & \textbf{49.6\%} & \textbf{58.6\%} & \textbf{0.026} & \textbf{0.004} & \textbf{0.040} \\
\bottomrule
\end{tabular}
}
\end{table}

\section{Weight Space Interpolation}
\label{sec:suppl_interpolation}

\begin{figure}[h]
    \centering
    \includegraphics[width=\linewidth]{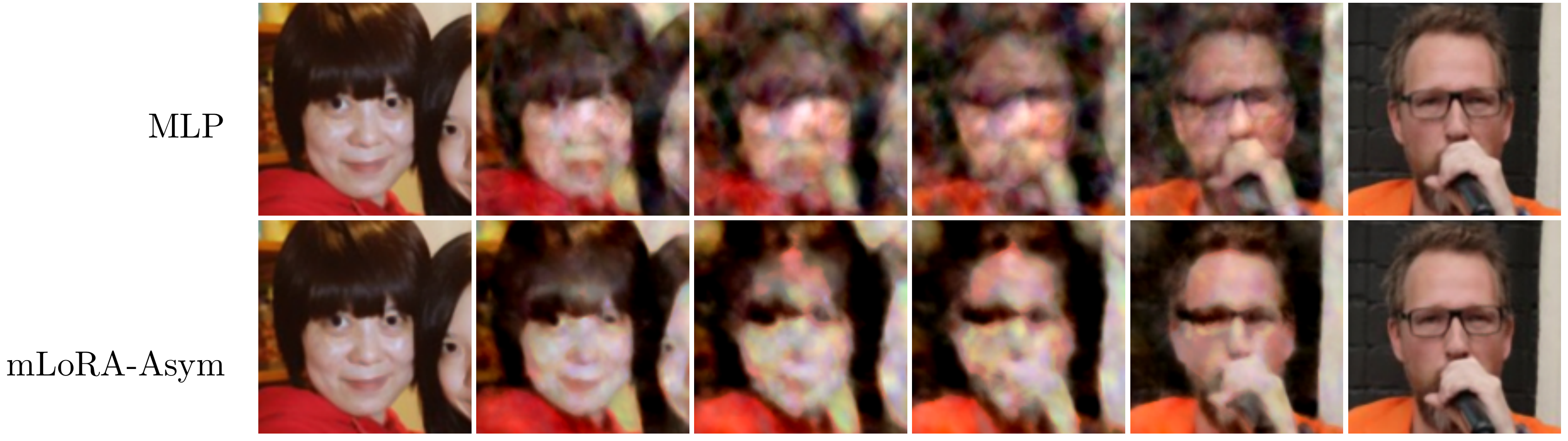}
    \caption{\textbf{Weight space interpolation.} Linear interpolation between pairs of mLoRA-Asym weight representations on FFHQ. Columns show the two endpoint instances (leftmost, rightmost) and intermediate interpolated reconstructions.}
    \label{fig:interp}
\end{figure}

Figure~\ref{fig:interp} visualizes linear interpolations between pairs of mLoRA weight representations on FFHQ. We linearly interpolate between two instances' LoRA weight pairs $(\mathbf{A}_1, \mathbf{B}_1)$ and $(\mathbf{A}_2, \mathbf{B}_2)$, evaluating the resulting neural field at each interpolation step. While the interpolated weights do not always produce smooth, perceptually gradual transitions between instances (as would be expected from a continuous learned latent space), this does not undermine our claims about the quality of the weight space representations. Smooth interpolation is characteristic of continuous latent spaces specifically optimized for this property, such as VAE latent codes, but structured representations like VQ-VAE quantized codes and point-cloud latents also lack this property yet achieve strong generative performance~\cite{vahdat2022lion}. Our experiments demonstrate that mLoRA weights support high-quality generation (Tables~2--3 in the main paper) and exhibit clear semantic structure for classification (Table~4, Figure~5 in the main paper), which are the primary criteria for effective weight space representations.

\section{Additional Qualitative Results}
\label{sec:suppl_qualitative}

We provide additional qualitative results on diffusion generation. Please see Figure~\ref{fig:qualitative_additional_airplane} for results on ShapeNet Airplanes, Figure~\ref{fig:qualitative_additional_multi} for results on ShapeNet Multi, and Figure~\ref{fig:qualitative_additional_ffhq} for results on FFHQ.

\begin{figure}[h]
    \centering
    \includegraphics[width=\linewidth]{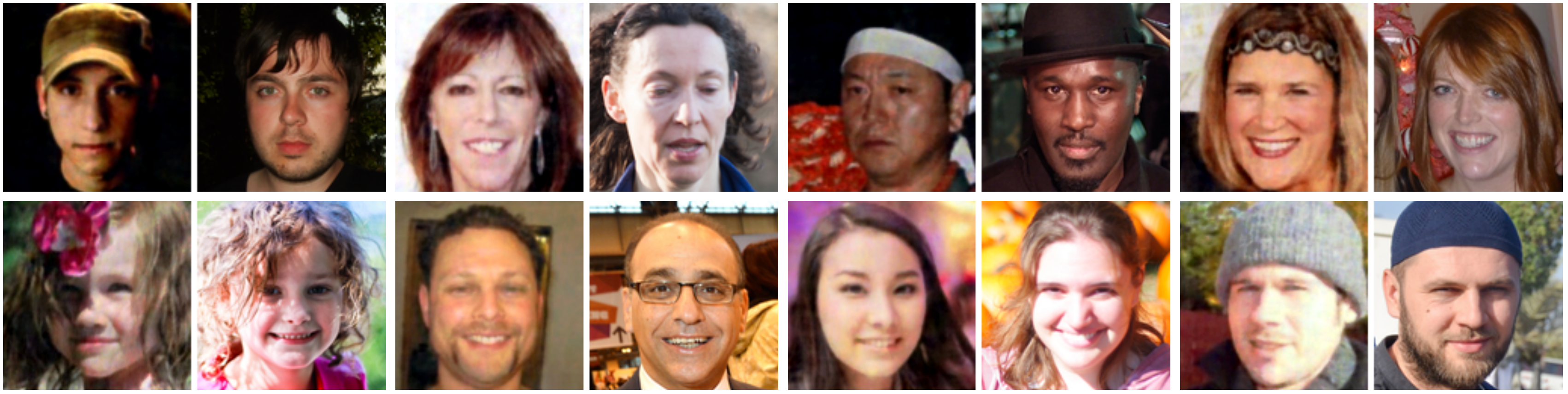}
    \caption{\textbf{Novelty check.} For each generated FFHQ image (left), we show its nearest neighbor from the training set retrieved by CLIP feature similarity (right). Generated samples are visually distinct from training data, confirming generalization rather than memorization.}
    \label{fig:novelty}
\end{figure}

\begin{figure*}
    \centering
    \includegraphics[width=0.95\linewidth]{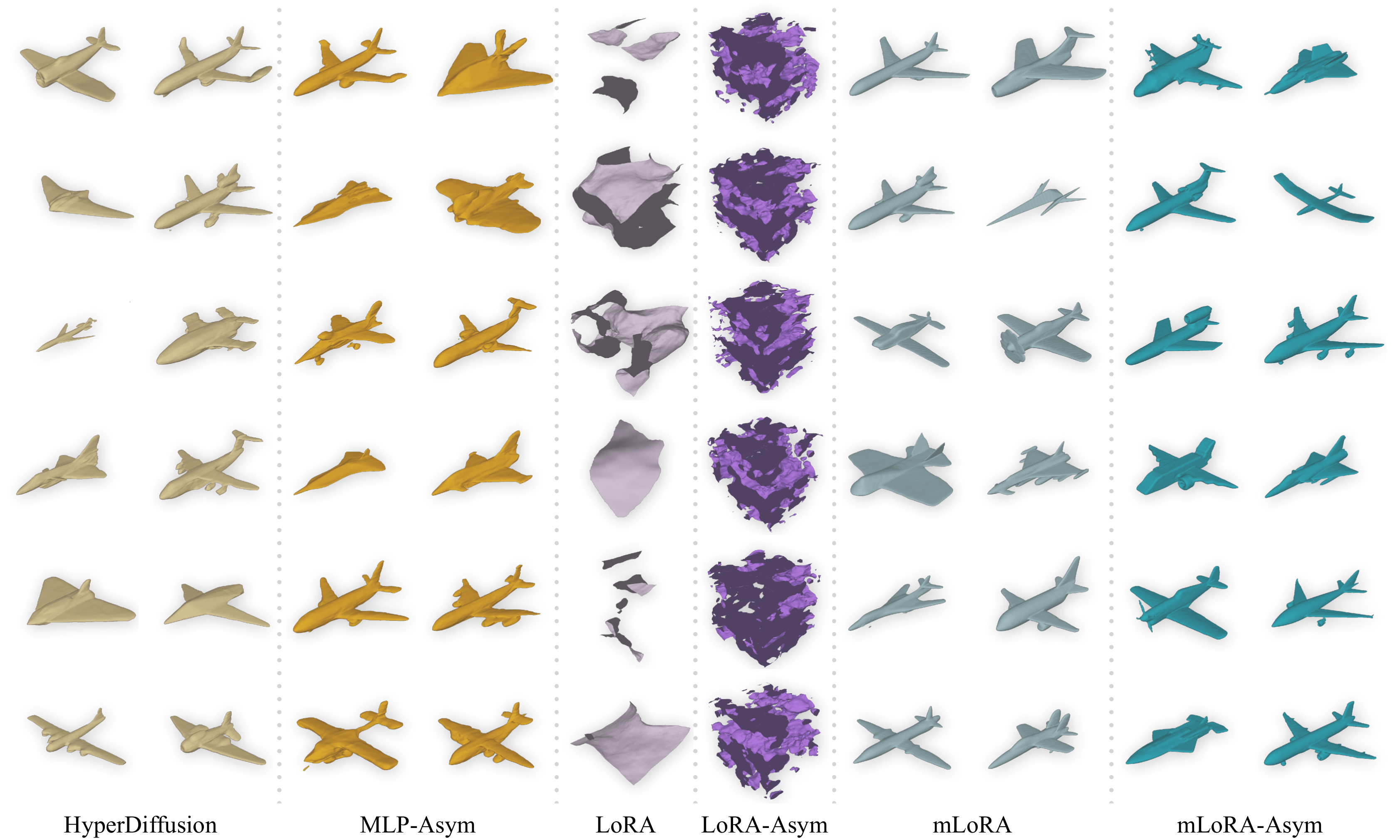}
    \caption{\textbf{Additional qualitative generation results on ShapeNet - Airplanes.}}
    \label{fig:qualitative_additional_airplane}
\end{figure*}

\begin{figure*}
    \centering
    \includegraphics[width=0.95\linewidth]{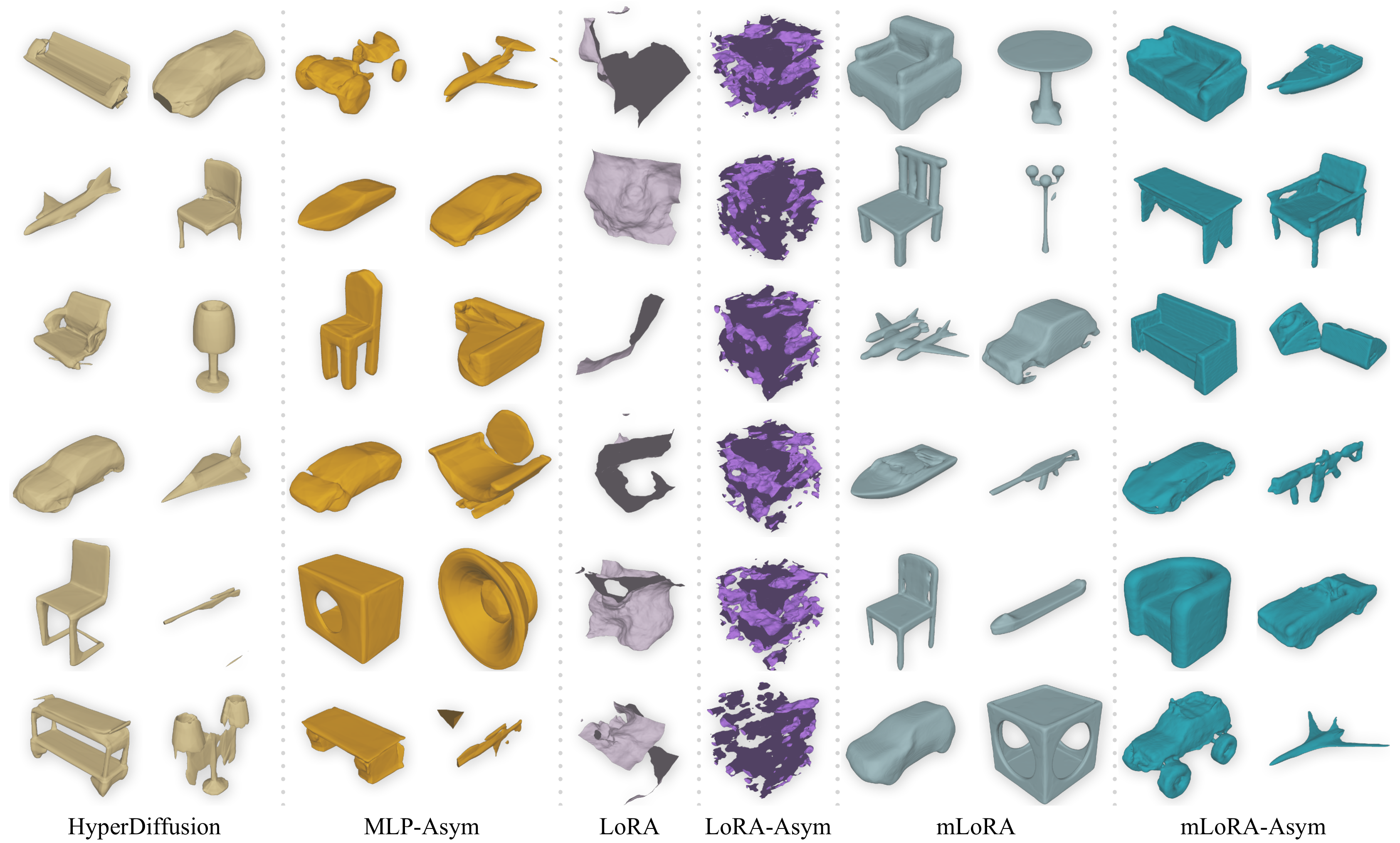}
    \caption{\textbf{Additional qualitative generation results on ShapeNet - Multi.}}
    \label{fig:qualitative_additional_multi}
\end{figure*}

\begin{figure*}
    \centering
    \includegraphics[width=0.95\linewidth]{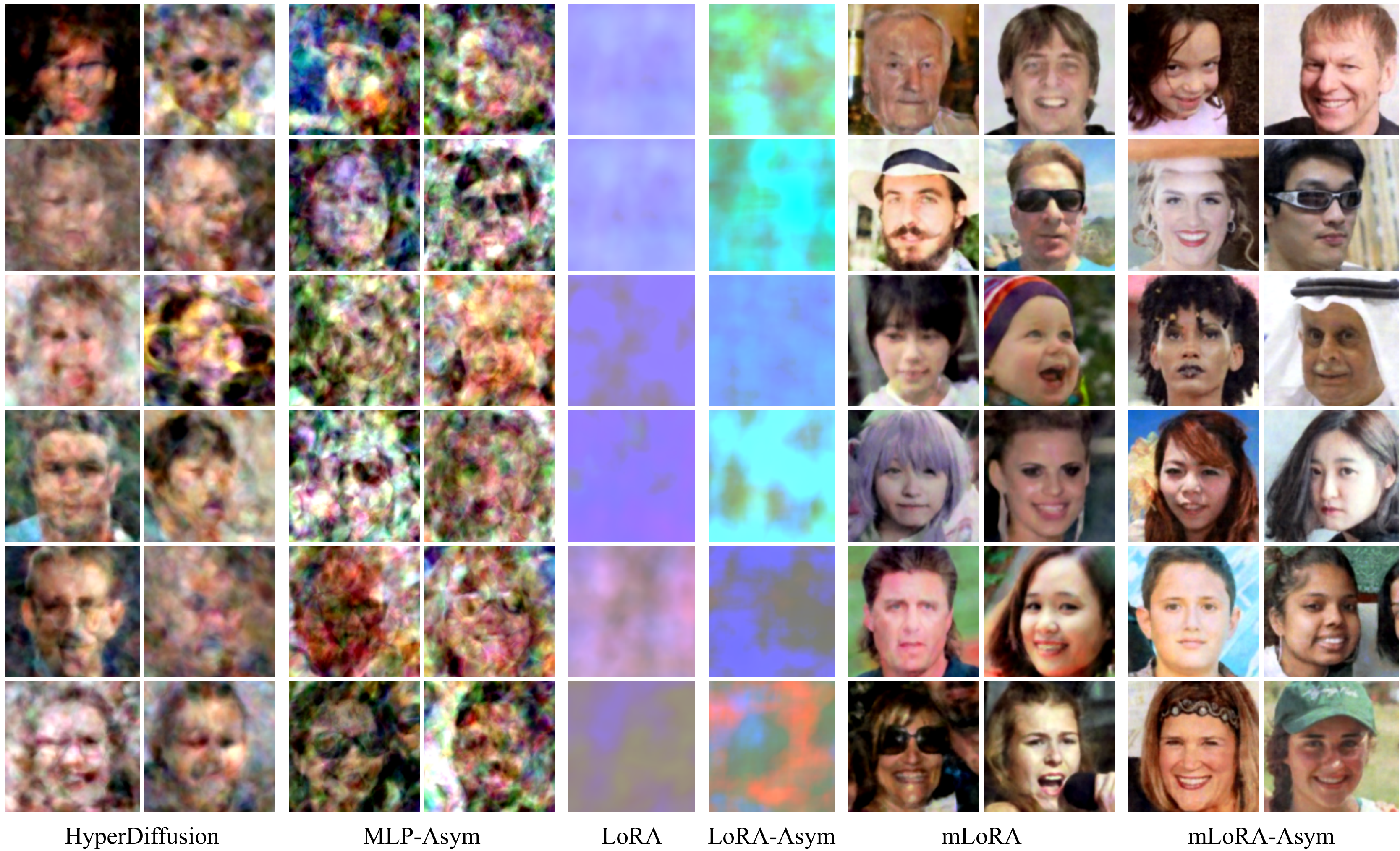}
    \caption{\textbf{Additional qualitative generation results on FFHQ.}}
    \label{fig:qualitative_additional_ffhq}
\end{figure*}

To verify that generated samples are novel rather than memorized reproductions of training data, we perform a CLIP-based nearest-neighbor analysis on FFHQ generations from the mLoRA-Asym configuration. For each generated image, we retrieve its nearest neighbor from the training set using CLIP feature similarity. Figure~\ref{fig:novelty} shows paired comparisons (left: generated, right: nearest training neighbor). The generated images are visually distinct from their nearest training neighbors, confirming that the diffusion model generalizes rather than memorizes.

\section{Limitation and Future Work}
\label{sec:suppl_limitation}

While our work establishes that neural network weights can serve as effective data representations, several limitations present opportunities for future research.

First, our approach requires all instances to share the same pre-trained base model and initialization. This is a practical limitation: the base model may not suit all INR architectures, and meaningful weight space comparisons between instances trained on different base models are not possible. Future work could explore methods to reduce this requirement or to align weight spaces across different base models.

Second, our approach requires finetuning a base-model which is computationally more expensive than fitting small MLPs. This requirement prevents evaluation on datasets with hundreds of thousands of samples, constraining our experiments to datasets with thousands of instances. Future work could explore methods to eliminate this requirement, or develop more computationally efficient multiplicative LoRA adaptation procedures.


Third, while our method achieves the first successful weight space generation on relatively high resolution natural images and demonstrates superior performance compared to prior weight space methods, the generation quality does not yet match state-of-the-art image generative models such as latent diffusion models. Future work could focus on closing this performance gap while preserving data modality agnosticism.

\end{document}